\documentclass[journal]{IEEEtran}
\usepackage{epsfig}
\usepackage{graphicx}
\usepackage{amsmath}
\usepackage{amssymb}
\usepackage{relsize}
\usepackage{caption,subcaption}
\usepackage{verbatim}
\usepackage{hyperref}

\usepackage{color}
\usepackage[table]{xcolor}

\definecolor{blue}{rgb}{0,0,0}  % blue used in the original response. For the decision response, this should stay {0,0,0}
\definecolor{bluedec}{rgb}{0,0,0}  % blue for the recision response; black for submission
\newcommand{\algname}{{VoxelMorph}} 
\DeclareMathOperator*{\argmin}{arg\,min}
\newcommand{\bphi}{\boldsymbol{\phi}}
\newcommand{\biphi}{\boldsymbol{\phi}}  %^{-1}

\newcommand{\bu}{\mathbf{u}}
\newcommand{\bp}{\mathbf{p}}
\newcommand{\bmoving}{m}
\newcommand{\bfixed}{f}
\newcommand{\bs}{s}

\usepackage{cite}
\ifCLASSINFOpdf
\else
\fi
\begin{document}
\title{VoxelMorph: A Learning Framework for Deformable Medical Image Registration}

\author{Guha Balakrishnan, \thanks{Guha Balakrishnan, Amy Zhao and John Guttag are with the Computer Science and Artificial Intelligence Lab, MIT}
        Amy Zhao,
        Mert R. Sabuncu, \thanks{Mert Sabuncu is with the the School of Electrical and Computer Engineering, and Meinig School of Biomedical Engineering, Cornell University.}
        John Guttag,
        and~Adrian~V.~Dalca \thanks{Adrian V. Dalca is with the Computer Science and Artificial Intelligence Lab, MIT and also Martinos Center for Biomedical Imaging, MGH, HMS.}}% <-this % stops a space

\maketitle

{\color{blue}\begin{abstract}
We present \algname{}, a fast learning-based {\color{bluedec}framework} for deformable, pairwise medical image registration. Traditional registration methods optimize an objective function for each pair of images, which can be time-consuming for large datasets or rich deformation models. In contrast to this approach, and building on recent learning-based methods, we formulate registration as a function that maps an input image pair to a deformation field that aligns these images. We parameterize the function via a convolutional neural network (CNN), and optimize the parameters of the neural network on a set of images. Given a new pair of scans, \algname{} rapidly computes a deformation field by directly evaluating the function. In this work, we explore two different training strategies. In the first (unsupervised) setting, we train the model to maximize standard image matching objective functions that are based on the image intensities. In the second setting, we leverage auxiliary segmentations available in the training data. We demonstrate that the unsupervised model's accuracy is comparable to state-of-the-art methods, while operating orders of magnitude faster. We also show that \algname{} trained with auxiliary data improves registration accuracy at test time, {\color{bluedec} and evaluate the effect of training set size on registration}. Our method promises to speed up medical image analysis and processing pipelines, while facilitating novel directions in learning-based registration and its applications. Our code is freely available at \url{http://voxelmorph.csail.mit.edu}.
\end{abstract}}


% \fi
\IEEEpeerreviewmaketitle

%\IEEEraisesectionheading{\section{Introduction}\label{sec:introduction}}
\vspace{-0.2cm}
\section{Introduction}\label{sec:introduction}
\IEEEPARstart{D}{eformable} registration is a fundamental task in a variety of medical imaging studies, and has been a topic of active research for decades. In deformable registration, a dense, non-linear correspondence is established between a pair of images, such as 3D {\color{blue}magnetic resonance (MR)} brain scans. Traditional registration methods solve an optimization problem for each volume pair by aligning voxels with similar appearance while enforcing constraints on the registration mapping. {\color{blue}Unfortunately, solving a pairwise optimization can be computationally intensive, and therefore slow in practice.  For example, state-of-the-art algorithms running on the CPU can require tens of minutes to hours to register a pair of scans with high accuracy~\cite{fischl2012,klein2009,avants2011}. Recent GPU implementations have reduced this runtime to just minutes, but require a GPU for each registration~\cite{modat2010fast}.}

{\color{blue}We present} a novel registration method that learns a parametrized registration \emph{function} from a collection of volumes. We implement the function using a convolutional neural network (CNN), that takes two $n$-D input volumes and outputs a mapping of all voxels of one volume to another volume. The parameters of the network, i.e.\ the convolutional kernel weights, {\color{bluedec}can be} optimized using {\color{blue}only} a training set of volumes from the dataset of interest. 
%{\color{bluedec} In addition, if auxiliary data, like manual segmentation maps, is available, these can optionally be used to also drive the network parameters. }
{\color{blue}The} procedure learns a common representation that enables alignment of a new pair of volumes from the same distribution. In essence, we replace a costly optimization {\color{blue}solved} for each test image pair with one global function optimization during a training phase. {\color{blue} Registration of a \textit{new} test scan pair is achieved by simply evaluating the learned function on the given volumes, resulting in rapid registration, even on a CPU.} We implement our method as a general purpose framework, \algname{}, available at \url{http://voxelmorph.csail.mit.edu}\footnote{\color{bluedec}We implement VoxelMorph as a flexible framework that includes the methods proposed in this manuscript, as well as extensions that are beyond the scope of this work~\cite{dalca2018}}.

{\color{blue}In the learning-based framework of VoxelMorph, we are free to adopt any differentiable objective function, and in this paper we present two possible choices. The first approach, which we refer to as unsupervised}\footnote{\color{bluedec} We use the term \textit{unsupervised} to underscore the fact that VoxelMorph is a learning method (with images as input and deformations as output) that requires no deformation fields during training. \color{blue}Alternatively, such methods have also been termed \textit{self-supervised}, to highlight the lack of supervision, or \textit{end-to-end}, to highlight that no external computation is necessary as part of a pipeline (such as computing 'true' deformation fields).}, uses only the input volume pair and the registration field computed by the model. Similar to traditional image registration algorithms, this loss function {\color{blue}quantifies the dissimilarity between the intensities of the two images and the spatial regularity of the deformation. The second approach also leverages} anatomical segmentations available at training time for a subset of the data, {\color{bluedec}to learn network parameters}.

Throughout this study, we use the example of registering 3D MR brain scans. However, our method is broadly applicable to other registration tasks, both within and beyond the medical imaging domain. We evaluate our work on a multi-study dataset of over 3,500 scans containing images of healthy and diseased brains from a variety of age groups. Our unsupervised model achieves comparable accuracy to state-of-the-art registration, while taking orders-of-magnitude less time. Registration with \algname{} requires less than a minute using a CPU and under a second on a GPU, {\color{blue}in contrast to the state-of-the-art baselines which take tens of minutes to over two hours on a CPU.} %This is of significant practical importance for many medical image tasks, including population analyses. %We also show that leveraging anatomical segmentation labels during training significantly improves registration accuracy while still retaining its rapid runtime. 

	This paper extends a preliminary version of the work
	presented at the 2018 International Conference on Computer
	Vision and Pattern Recognition~\cite{balakrishnan2018}. {\color{bluedec} We build on that work by expanding analyses, and introducing an auxiliary learning model that can use anatomical segmentations during training to improve registration on new test image pairs for which segmentation maps are not available.} {\color{bluedec} 
	We focus on providing a thorough analysis of the behavior of the VoxelMorph algorithm
	using two loss functions and a variety of settings, as follows. We test the unsupervised approach on more datasets and both atlas-based and subject-to-subject registration. We then explore cases where different types and numbers of anatomical region segmentations are available during
	training as auxiliary information, and evaluate the effect on registration of test data where segmentations are not available. We present an empirical analysis quantifying
	the effect of training set size on accuracy, and show how
	instance-specific optimization can improve results. Finally, we perform sensitivity analyses with respect to the hyperparameter choices, and discuss an interpretation of our model as amortized optimization.
}

The paper is organized as follows. Section 2 introduces medical image registration and Section 3 describes related work. Section 4 presents our methods. Section 5 presents experimental results on MRI data. We discuss insights of the results and conclude in Section 6.
\vspace{-0.25cm}
\section{Background}
\begin{figure}[t!]
	\begin{center}
		\includegraphics[width=\linewidth]{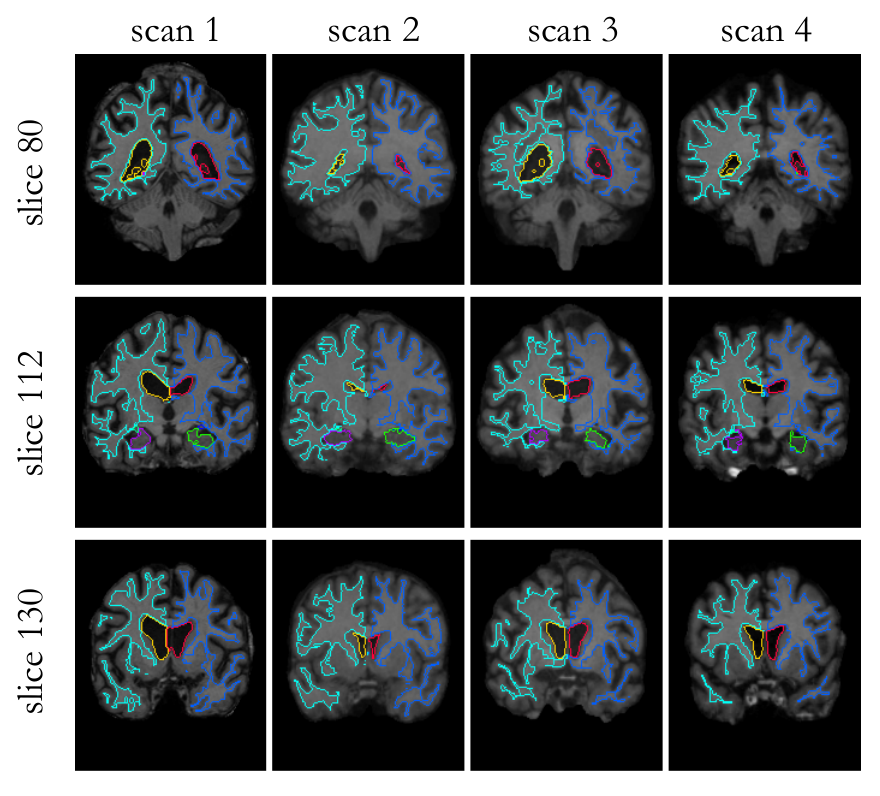}
	\end{center}
	\vspace{-0.3cm}
	\caption{Example coronal slices from the MRI brain dataset, after affine alignment. Each column is a different scan (subject) and each row is a different coronal slice. Some anatomical regions are outlined using different colors: L/R white matter in light/dark blue, L/R ventricles in yellow/red, and L/R hippocampi in purple/green. There are significant structural differences across scans, necessitating a deformable registration step to analyze inter-scan variations. \vspace{-0.4cm}}
	\label{fig:examples}
\end{figure}

In the traditional volume registration formulation, one (moving or source) volume is warped to align with a second (fixed or target) volume. Fig.~\ref{fig:examples} shows sample 2D coronal slices taken from 3D MRI volumes, with boundaries of several anatomical structures outlined. There is significant variability across subjects, caused by natural anatomical brain variations and differences in health state. Deformable registration enables comparison of structures between scans. Such analyses are useful for understanding variability across populations or the evolution of brain anatomy over time for individuals with disease. Deformable registration strategies often involve two steps: an initial affine transformation for global alignment, followed by a much slower deformable transformation with more degrees of freedom. We concentrate on the latter step, in which we compute a dense, nonlinear correspondence for all voxels.

Most existing deformable registration algorithms iteratively optimize a transformation based on an energy function~\cite{sotiras2013deformable}. Let $\bfixed$ and $\bmoving$ denote the fixed and moving images, respectively, and let $\bphi$ {\color{bluedec}be the registration field that maps coordinates of $\bfixed$ to coordinates of $\bmoving$}. The optimization problem can be written as:
\begin{align}
\label{eqn:energy}
\widehat{\bphi} &= \argmin_{\bphi}\mathcal{L}(\bfixed,\bmoving,\bphi) \\ 
&=  \argmin_{\bphi} \mathcal{L}_{sim}(\bfixed,\bmoving \circ \biphi) + \lambda \mathcal{L}_{smooth}(\bphi), 
\end{align}
where $\bmoving \circ \biphi $ represents $\bmoving$ warped by $\bphi$, function $\mathcal{L}_{sim}(\cdot,\cdot)$ measures image similarity between its two inputs, $\mathcal{L}_{smooth}(\cdot)$ imposes regularization, and $\lambda$ is the regularization trade-off parameter. 

There are several common formulations for $\bphi$, $\mathcal{L}_{sim}$ and $\mathcal{L}_{smooth}$. Often, {\color{blue}$\bphi$ is characterized by a displacement vector field~$\bu$ specifying the vector offset from $\bfixed$ to $\bmoving$ for each voxel:~$\bphi = Id + \bu$, where~$Id$ is the identity transform~\cite{bajcsy1989}}. {\color{bluedec} Diffeomorphic transforms model $\bphi$ through the integral of a velocity vector field, preserving topology and maintaining invertibility on the transformation~\cite{beg2005}}. Common metrics used for $\mathcal{L}_{sim}$ include intensity mean squared error, mutual information~\cite{viola1997alignment}, and cross-correlation~\cite{avants2008}. The latter two are particularly useful when volumes have varying intensity distributions and contrasts. $\mathcal{L}_{smooth}$ enforces a spatially smooth deformation, often modeled {\color{bluedec}as a function of the spatial gradients of~$\bu$}. 

Traditional algorithms optimize~\eqref{eqn:energy} for each volume pair. This is expensive when registering many volumes, for example as part of population-wide analyses. In contrast, we assume that a field can be computed by a parameterized function of the data. We optimize the function parameters by minimizing the expected energy of the form of~\eqref{eqn:energy} over a dataset of volume pairs. Essentially, we replace pair-specific optimization of the deformation field by global optimization of the shared parameters, which {\color{blue}in other domains} has been referred to as amortization~\cite{kim2018semi, sonderby2016, zhang2017advances, cremer2018inference}. Once the global function is estimated, a field can be produced by evaluating the function on a given volume pair. In this paper, we use a displacement-based vector field {\color{blue}representation}, and focus on various aspects of the learning framework and its advantages. However, we recently demonstrated that velocity-based representations are also possible in a \algname{}-like framework, {\color{bluedec} also included in our codebase}~\cite{dalca2018}.

\vspace{-0.15cm}
\section{Related Work}

\subsection{Medical Image Registration (Non-learning-based)}
There is extensive work in 3D medical image registration~\cite{ashburner2007,avants2008,bajcsy1989,beg2005,dalca2016,glocker2008,thirion1998, yeo2010learning, zhang2017}. Several studies optimize within the space of displacement vector fields. These include elastic-type models~\cite{bajcsy1989, davatzikos1997, shen2002}, statistical parametric mapping~\cite{ashburner2000}, free-form deformations with b-splines~\cite{rueckert1999}, discrete methods~\cite{dalca2016, glocker2008} and Demons~\cite{pennec1999, thirion1998}. Diffeomorphic transforms, which are topology-preserving, have shown remarkable success in various computational anatomy studies. Popular formulations include Large Diffeomorphic Distance Metric Mapping (LDDMM)~\cite{beg2005, cao2005large, ceritoglu2009multi, hernandez2009registration, joshi2000landmark, miller2005increasing, oishi2009atlas, zhang2017}, DARTEL~\cite{ashburner2007}, diffeomorphic demons~\cite{vercauteren2009}, and standard symmetric normalization (SyN)~\cite{avants2008}. All of these non-learning-based approaches optimize an energy function for each image pair, resulting in slow registration. {\color{blue} Recent GPU-based algorithms build on these concepts to reduce algorithm runtime to several minutes, but require a GPU to be available for each registration~\cite{modat2010fast,modat2014global}}.

\subsection{Medical Image Registration (Learning-based)}
There are several recent papers proposing neural networks to learn a function for medical image registration. Most of these rely on ground truth warp fields~\cite{cao2017deformable,krebs2017,rohe2017,sokooti2017,yang2017}, {\color{blue} which are either obtained by simulating deformations and deformed images, or running classical registration methods on pairs of scans. Some also use image similarity to help guide the registration~\cite{cao2017deformable}.} {\color{blue} While supervised methods present a promising direction, ground
	truth warp fields derived via conventional registration tools as ground truth can be cumbersome to acquire and can restrict the type of deformations
	that are learned. 
	%For example, if the conventional tool used is suboptimal for a particular task, the learning algorithm is similarly likely to perform suboptimally.
	} 
	{\color{blue}In contrast, VoxelMorph is unsupervised, and is also capable of leveraging auxiliary information such as segmentations during training if those are available. 

Two recent papers~\cite{devos2017,li2017}, were the first to present unsupervised learning based image registration methods. Both propose a neural network consisting of a CNN and spatial transformation function~\cite{jaderberg2015} that warps images to one another. However, these two initial methods are only demonstrated on limited subsets of volumes, such as 3D subregions~\cite{li2017} or 2D slices~\cite{devos2017}, and support only small transformations~\cite{devos2017}. %{\color{red}Others~\cite{devos2017} employ regularization implicitly determined by interpolation methods.} 
}

{\color{blue} A recent method has proposed a segmentation driven cost function to be used in registering different imaging modalities -- T2w MRI and 3D ultrasound -- within the same subject~\cite{hu2018weakly,hu2018label}. The authors demonstrate that a loss functions based solely on segmentation maps can lead to an accurate within-subject cross-modality registration network. Parallel to this work, in one of our experiments, we demonstrate the use of segmentation maps during training in subject-to-atlas registration. We provide an analysis of the effect of different anatomical label availability on overall registration quality, and evaluate how a combination of segmentation and image based losses behaves in various scenarios. {\color{bluedec} 
		We find that a segmentation-based loss can be helpful, for example if the input segment labels are the same as those we evaluate on (consistent with~\cite{hu2018weakly}, and~\cite{hu2018label}). We also show that the image-based and smoothness losses are still necessary, especially when we evaluate registration accuracy on labels not observed during training, and to encourage deformation regularity.
	}
}

%In contrast to these methods, our generalizable method is demonstrated on entire 3D volumes between different subjects, handles large deformations, and supports any differentiable cost function. 
%We extend the preliminary version of our method~\cite{balakrishnan2018} to show how to leverage auxiliary information during training, and provide extensive analysis of the effect of different label subsets on registration quality.

\vspace{-0.15cm}
\subsection{2D Image Alignment}
Optical flow estimation is a related registration problem for 2D images. Optical flow algorithms return a dense displacement vector field depicting small displacements between a pair of 2D images. Traditional optical flow approaches typically solve an optimization problem similar to~\eqref{eqn:energy} using variational methods~\cite{brox2004,horn1980,sun2010}. Extensions that better handle large displacements or dramatic changes in appearance include feature-based matching~\cite{brox2011,liu2011} and nearest neighbor fields~\cite{chen2013}. 

{\color{bluedec}In recent years, several learning-based approaches to optical flow estimation using neural networks have been proposed~\cite{ahmadi2016unsupervised, dosovitskiy2015, ilg2017flownet, jason2016back, ranjan2017optical, tran2016deep}. These algorithms take a pair of images as input, and use a convolutional neural network to learn image features that capture the concept of optical flow from data. Several of these works require supervision in the form of ground truth flow fields~\cite{dosovitskiy2015, ilg2017flownet, ranjan2017optical, tran2016deep}, while we build on a few that use an unsupervised objective~\cite{ahmadi2016unsupervised, jason2016back}. The spatial transform layer enables neural networks to perform both global parametric 2D image alignment~\cite{jaderberg2015} and dense spatial transformations~\cite{jason2016back, park2017,zhou2016} without requiring supervised labels. An alternative approach to dense estimation is to use CNNs to match image patches~\cite{bailer2017cnn, gadot2016patchbatch, thewlis2016fully, weinzaepfel2013}. These methods require exhaustive matching of patches, resulting in slow runtime.

We build on these ideas and extend the spatial transformer to achieve n-D volume registration, and further show how leveraging image segmentations during training can improve registration accuracy at test time.}

\begin{figure*}[h!]
\begin{center}
\includegraphics[scale=0.7]{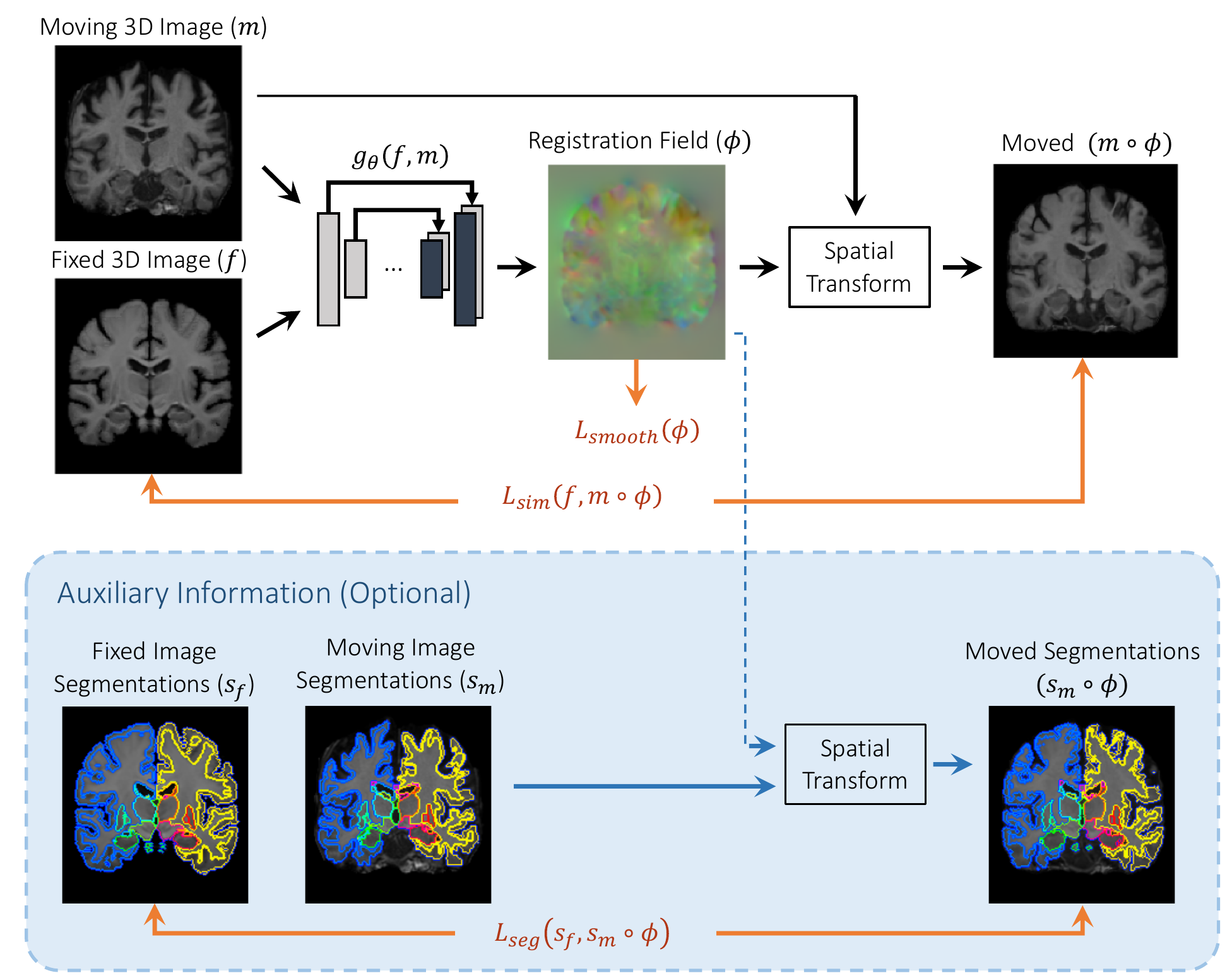}
\end{center}
\vspace{-0.2cm}
\caption{Overview of the method. We learn parameters $\theta$ for a function $g_{\theta}(\cdot, \cdot)$, and register 3D volume $\bmoving$ to a second, fixed volume $\bfixed$. During training, we warp $\bmoving$ with $\bphi$ using a spatial transformer function. Optionally, auxiliary information such as anatomical segmentations $\mathbf{s_f}, \mathbf{s_m}$ can be leveraged during training (blue box).\vspace{-0.2cm}}
\label{fig:overview}
\end{figure*}

\vspace{-0.2cm}
\section{Method}
Let $\bfixed,\bmoving$ be two image volumes defined over an $n$-D spatial domain $\Omega \subset \mathbb{R}^n$. For the rest of this paper, we focus on the case $n=3$ but our method and implementation are dimension independent. For simplicity we assume that $\bfixed$ and $\bmoving$ contain single-channel, grayscale data. We also assume that $\bfixed$ and $\bmoving$ are affinely aligned as a preprocessing step, so that the only source of misalignment between the volumes is nonlinear. Many packages are available for rapid affine alignment.

{\color{bluedec}We model a function $g_{\theta}(\bfixed,\bmoving) = \bu$ using a convolutional neural network (CNN), where $\theta$ are network parameters, the kernels of the convolutional layers. The displacement field $\bu$ between~$\bfixed$ and~$\bmoving$ is in practice stored in a ${n+1}$-dimensional image. That is, for each voxel~$\mathbf{p} \in \Omega$, $\bu (\mathbf{p})$ is a displacement such that $\bfixed(\bp)$ and $[\bmoving \circ \biphi ](\bp) $ correspond to similar anatomical locations, where the map $\bphi = Id + \bu$ is formed using an identity transform and~$\bu$.} 

Fig.~\ref{fig:overview} presents an overview of our method. The network takes $\bfixed$ and $\bmoving$ as input, and computes $\bphi$ using a set of parameters~$\theta$. We warp~$\bmoving$ to~$\bmoving \circ \biphi$ using a spatial transformation function, enabling evaluation of the similarity of~$\bmoving \circ \biphi$ and $\bfixed$. {\color{blue}Given unseen images $\bfixed$ and $\bmoving$ during test time, we obtain a registration field by evaluating $g_{\theta}(\bfixed,\bmoving)$.}

{\color{blue}We use (single-element) stochastic gradient descent to find optimal parameters $\hat{\theta}$ by minimizing an expected loss function using a training dataset}. We propose two unsupervised loss functions in this work. The first captures image similarity and field smoothness, while the second also leverages anatomical segmentations.  We describe our CNN architecture and the two loss functions in detail in the next sections. 

\begin{comment}
 of the form~\eqref{eqn:energy}:

\begin{align}
\hat{\theta} &= \argmin_{\theta} \mathbb{E}_{(\bfixed,\bmoving)\sim D}\left[\mathcal{L}\left(\bfixed,\bmoving,g_{\theta}(\bfixed,\bmoving)\right)\right], 
\label{eqn:poploss}
\end{align}

\noindent where $D$ is the dataset distribution. \end{comment}

\vspace{-0.2cm}
\subsection{\algname{}  CNN Architecture}
In this section we describe the particular architecture used in our experiments, but emphasize that a wide range of architectures may work similarly well and that the exact architecture is not our focus. The parametrization of $g_{\theta} (\cdot, \cdot)$ is based on a convolutional neural network architecture similar to UNet~\cite{isola2017,ronneberger2015}, which consists of encoder and decoder sections with skip connections. 

%\begin{figure}[h!]
%\begin{center}
%\includegraphics[width=\linewidth]{figures/unets_scalesbottom.pdf}
%\end{center}
%\vspace{-0.2cm}
%\caption{Convolutional UNet architecture implementing $g_\theta(\bfixed,\bmoving)$. Each rectangle represents a 3D volume, generated from its preceding volume using a 3D convolutional network layer. The spatial resolution of each volume with respect to the input volume is printed underneath, and the number of channels of each volume is printed within. The encoder uses convolutions with a stride of 2, while in the decoder, each convolution is followed by an upsampling layer. Arrows represent skip connections, which concatenate encoder and decoder features. The full-resolution volume is further refined using several convolutions.}
%\label{fig:unets}
%\end{figure}

\begin{figure}[h!]
	\begin{center}
		\includegraphics[width=1\linewidth]{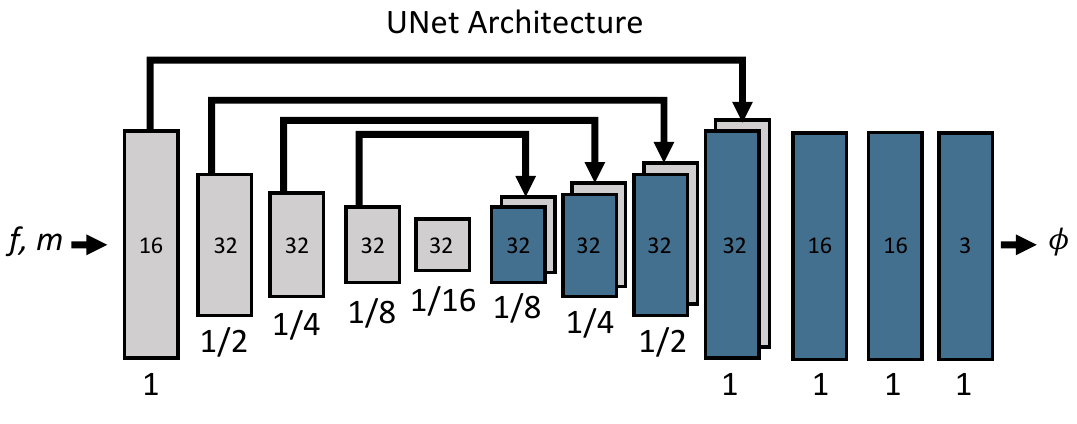}
	\end{center}
	\caption{\color{blue}Convolutional UNet architecture implementing $g_\theta(\bfixed,\bmoving)$. Each rectangle represents a 3D volume, generated from the preceding volume using a 3D convolutional network layer. The spatial resolution of each volume with respect to the input volume is printed underneath. 
		%The encoder uses convolutions with 16, 32, 32, 32 filters respectively, each with a stride of 2. 
		In the decoder, we use several 32-filter convolutions, each followed by an upsampling layer, to bring the volume back to full resolution. Arrows represent skip connections, which concatenate encoder and decoder features. The full-resolution volume is further refined using several convolutions.
		% with 32, 16, 16 filters respectively. 
	}
	\label{fig:unets}
\end{figure}

{\color{blue} Fig.~\ref{fig:unets} depicts the network used in VoxelMorph, which takes a single input formed by concatenating $\bmoving$ and $\bfixed$ into a 2-channel 3D image. In our experiments, the input is of size $160\times192\times224\times2$, but the framework is not limited by a particular size. We apply 3D convolutions in both the encoder and decoder stages using a kernel size of $3$, and a stride of 2. Each convolution is followed by a LeakyReLU layer with parameter $0.2$. The convolutional layers capture hierarchical features of the input image pair, used to estimate $\bphi$. In the encoder, we use strided convolutions to reduce the spatial dimensions in half at each layer. Successive layers of the encoder therefore operate over coarser representations of the input, similar to the image pyramid used in traditional image registration work.}

In the decoding stage, we alternate between upsampling, convolutions and concatenating skip connections that propagate features learned during the encoding stages directly to layers generating the registration. Successive layers of the decoder operate on finer spatial scales, enabling precise anatomical alignment. The receptive fields of the convolutional kernels of the smallest layer should be at least as large as the maximum expected displacement between corresponding voxels in $\bfixed$ and $\bmoving$. In our architecture, the smallest layer applies convolutions over a volume $(1/16)^3$ of the size of the input images. %We provide further architecture details in the supplementary material.

\vspace{-0.2cm}
\subsection{Spatial Transformation Function}
The proposed method learns optimal parameter values in part by minimizing differences between $\bmoving \circ \biphi$ and $\bfixed$. In order to use standard gradient-based methods, we construct a differentiable operation based on spatial transformer networks~\cite{jaderberg2015} to compute $\bmoving \circ \biphi$. 

For each voxel $\bp$, we compute a (subpixel) voxel location $\bp' = \bp + \bu(\bp)$ in $\bmoving$. Because image values are only defined at integer locations, we linearly interpolate the values at the eight neighboring voxels:
\begin{equation}
\bmoving \circ \biphi(\bp) =\hspace{-0.25cm}\sum_{\mathbf{q} \in \mathcal{Z}(\bp')}\hspace{-0.25cm}\bmoving(\mathbf{q})\hspace{-0.25cm}\prod_{d \in \{x,y,z\}}\hspace{-0.25cm}(1-|\bp'_d - \mathbf{q}_d|),\\
\end{equation}
\noindent where $\mathcal{Z}(\bp')$ are the voxel neighbors of $\bp'$, and $d$ iterates over dimensions of $\Omega$. {\color{blue}Because we can compute gradients or subgradients},\footnote{{\color{blue}The absolute value is implemented with a subgradient of 0 at 0.}} we can backpropagate errors during optimization.  

\subsection{Loss Functions}
In this section, we propose two loss functions: an unsupervised loss $\mathcal{L}_{us}$ that evaluates the model using only the input volumes and generated registration field, and an auxiliary loss $\mathcal{L}_{a}$ that also leverages anatomical segmentations at training time. %In general, \algname{} can be trained with any differentiable loss function. 

\subsubsection{Unsupervised Loss Function}
The unsupervised loss $\mathcal{L}_{us}(\cdot, \cdot, \cdot)$ consists of two components: $\mathcal{L}_{sim}$ that penalizes differences in appearance, and $\mathcal{L}_{smooth}$ that penalizes local spatial variations in $\bphi$:
\begin{align}
\mathcal{L}_{us}(\bfixed,\bmoving,\bphi) = \mathcal{L}_{sim}(\bfixed,\bmoving \circ \biphi)  + \lambda \mathcal{L}_{smooth}(\bphi),
\label{eqn:unsuploss}
\end{align}
where $\lambda$ is a regularization parameter. We experimented with two often-used functions for $\mathcal{L}_{sim}$. The first is the mean squared voxelwise difference, applicable when $\bfixed$ and $\bmoving$ have similar image intensity distributions and local contrast:
\begin{align}
\color{bluedec}
%MSE(\bfixed,\bmoving \circ \biphi) = \frac{1}{|\Omega|}\sum_{p \in \Omega} \| \bfixed_{\bp} - [\bmoving \circ \biphi]_{\bp}\|^2.
MSE(\bfixed,\bmoving \circ \biphi) = \frac{1}{|\Omega|}\sum_{p \in \Omega} \left[ \bfixed (\bp) - [\bmoving \circ \biphi] (\bp) \right]^2.
\end{align}
The second is the local cross-correlation of $\bfixed$ and \mbox{$\bmoving \circ \biphi$}, which is more robust to intensity variations found across scans and datasets~\cite{avants2008}. Let $\hat{\bfixed}({\bp})$ and \mbox{$[ \hat{\bmoving} \circ \biphi ]({\bp})$} denote local mean intensity images:~\mbox{$\hat{\bfixed}({\bp}) = \frac{1}{n^3}\sum_{\bp_i}\bfixed (\bp_i)$}, where $\bp_i$ iterates over a $n^3$ volume around $\bp$, with $n=9$ in our experiments. The local cross-correlation of $\bfixed$ and $\bmoving \circ \biphi$ is written as:
\footnotesize
\begin{align}
&CC(\bfixed,\bmoving\circ \biphi)= \nonumber \\
&\sum\limits_{\bp \in \Omega} \frac{\left(\sum\limits_{\bp_i} (\bfixed ({\bp_i}) - \hat{\bfixed} ({\bp}))([\bmoving \circ \biphi] ({\bp_i}) - [\hat{\bmoving} \circ \biphi] ({\bp}))\right)^2}
{\left(\sum\limits_{\bp_i} (\bfixed ({\bp_i}) - \hat{\bfixed} ({\bp}))^2 \right)\left(\sum\limits_{\bp_i} ([\bmoving\circ \biphi] ({\bp_i}) - [\hat{\bmoving}\circ \biphi] ({\bp}))^2\right)}.
\end{align}
\normalsize
A higher CC indicates a better alignment, yielding the loss function: $\mathcal{L}_{sim}(\bfixed,\bmoving,\bphi) = - CC(\bfixed,\bmoving \circ \biphi)$. 

Minimizing $\mathcal{L}_{sim}$ will encourage $\bmoving \circ \biphi$ to approximate~$\bfixed$, but may generate a non-smooth $\bphi$ that is not physically realistic. We encourage a smooth displacement field $\bphi$ using a diffusion regularizer on the spatial gradients of displacement~$\bu$:
\begin{align}
\mathcal{L}_{smooth}(\bphi) = \sum_{\bp \in \Omega} \lVert \nabla {\color{blue}\bu(\bp)} \rVert ^2,
\end{align}
and approximate spatial gradients using differences between neighboring voxels. 
{\color{bluedec} Specifically, for \small \mbox{$\nabla \bu(\bp) = \left(\frac{\partial \bu(\bp)}{\partial x}, \frac{\partial \bu(\bp)}{\partial y}, \frac{\partial \bu(\bp)}{\partial z}\right)$}, \normalsize we approximate  \small $\frac{\partial \bu(\bp)}{\partial x} \approx \bu( (p_x + 1, p_y, p_z)) - \bu( (p_x, p_y, p_z))$, \normalsize and use similar approximations for~\small $\frac{\partial \bu(\bp)}{\partial y}$ and~$\frac{\partial \bu(\bp)}{\partial z}$.}\normalsize

\subsubsection{Auxiliary Data Loss Function}
Here, we describe how \algname{} can leverage auxiliary information available during training but not during testing. % The notion of a ground truth registration field is ill-defined, and using fields derived via conventional registration tools as ground truth can introduce a bias. Instead, we sometimes have 
Anatomical segmentation maps are sometimes available during training, and can be annotated by human experts or automated algorithms. A segmentation map assigns each voxel to an anatomical structure. If a registration field $\bphi$ represents accurate anatomical correspondences, the regions in $\bfixed$ and $\bmoving\circ \biphi$ corresponding to the same anatomical structure should overlap well. 

Let $\bs_{f}^k,\bs^k_m \circ \biphi$ be the voxels of structure $k$ for $\bfixed$ and $\bmoving \circ \biphi$, respectively. We quantify the volume overlap for structure $k$ using the Dice score~\cite{dice1945}:
\begin{equation}
\text{Dice}(\bs_f^k,\bs^k_m \circ \biphi) = 2 \cdot \frac{|\bs^k_f \cap (\bs^k_m \circ \biphi)|}{|\bs^k_f| + |\bs^k_m \circ \biphi|}.
\label{eqn:dice}
\end{equation}
A Dice score of $1$ indicates that the anatomy matches perfectly, and a score of 0 indicates that there is no overlap. We define the segmentation loss $\mathcal{L}_{seg}$ over all structures $k \in [1,K]$ as:
\begin{equation}
\mathcal{L}_{seg}(\bs_f,\bs_m \circ \biphi) = - \frac{1}{K} \sum_{k=1}^K \text{Dice}(\bs^k_f,\bs^k_m \circ \biphi) .
\label{eqn:sup}
\end{equation}
$\mathcal{L}_{seg}$ alone does not encourage smoothness and agreement of image appearance, which are essential to good registration. We therefore combine $\mathcal{L}_{seg}$ with~\eqref{eqn:unsuploss} to obtain the objective:
\begin{align}
&\mathcal{L}_{a}(\bfixed,\bmoving,\bs_f,\bs_m,\bphi) = \nonumber \\ 
&\mathcal{L}_{us}(\bfixed,\bmoving,\bphi) + \gamma \mathcal{L}_{seg}(\bs_f,\bs_m \circ \biphi),
\label{eqn:semi}
\end{align}
where $\gamma$ is a regularization parameter. 
%We emphasize that although the segmentations are used during training, they are not inputs to $g_\theta (\cdot, \cdot)$, as shown in Fig.~\ref{fig:overview}. We do not use them to register a new image pair during test time. 

{\color{blue} In our experiments, which use affinely aligned images, we demonstrate that loss~\eqref{eqn:semi} can lead to significant improvements. In general, and depending on the task, this loss can also be computed in a multiscale fashion as introduced in~\cite{hu2018weakly}, depending on quality of the initial alignment.}

Since anatomical labels are categorical, a naive implementation of linear interpolation to compute $\bs_m \circ \biphi$ is inappropriate, and a direct implementation of~\eqref{eqn:dice} might not be amenable to auto-differentiation frameworks. We design $\bs_f$ and $\bs_m$ to be image volumes with $K$ channels, where each channel is a binary mask specifying the spatial domain of a particular structure. {\color{blue}We compute $\bs_m \circ \biphi$ by spatially transforming each channel of $\bs_m$ using linear interpolation.} We then compute the numerator and denominator of~\eqref{eqn:dice} by multiplying and adding $\bs_f$ and $\bs_m \circ \biphi$, respectively.

\subsection{Amortized Optimization Interpretation}
Our method substitutes the pair-specific optimization over the deformation field $\bphi$ with a global optimization of function parameters $\theta$ for function $g_\theta(\cdot, \cdot)$. This process is sometimes referred to as amortized optimization~\cite{marino2018iterative}. Because the function $g_\theta(\cdot, \cdot)$ is tasked with estimating registration between any two images, the fact that parameters~$\theta$ are shared globally acts as a natural regularization. We demonstrate this aspect in Section~\ref{sec:reganalysis} (Regularization Analysis). In addition, the quality and generalizability of the deformations outputted by the function will depend on the data it is trained on. Indeed, the resulting deformation can be interpreted as simply an approximation or initialization to the optimal deformation $\bphi^*$, and the resulting difference is sometimes referred to as the amortization gap~\cite{cremer2018inference,marino2018iterative}. If desired, this initial deformation field could be improved using any instance-specific optimization. {\color{bluedec}In our experiments, we accomplish this by treating the
	resulting displacement~$\bu$ as model parameters, and fine-tuning
	the deformation for each particular scan independently using
	gradient descent. Essentially, this implements an auto-differentiation version of conventional registration, using VoxelMorph output as initialization.} However, most often we find that the \textit{initial} deformation, the \algname{} output, is already as accurate as state of the art results. We explore these aspects in experiments presented in Section~\ref{sec:amortized}.

\begin{figure}[tb!]
	\begin{center}
		\includegraphics[width=\linewidth]{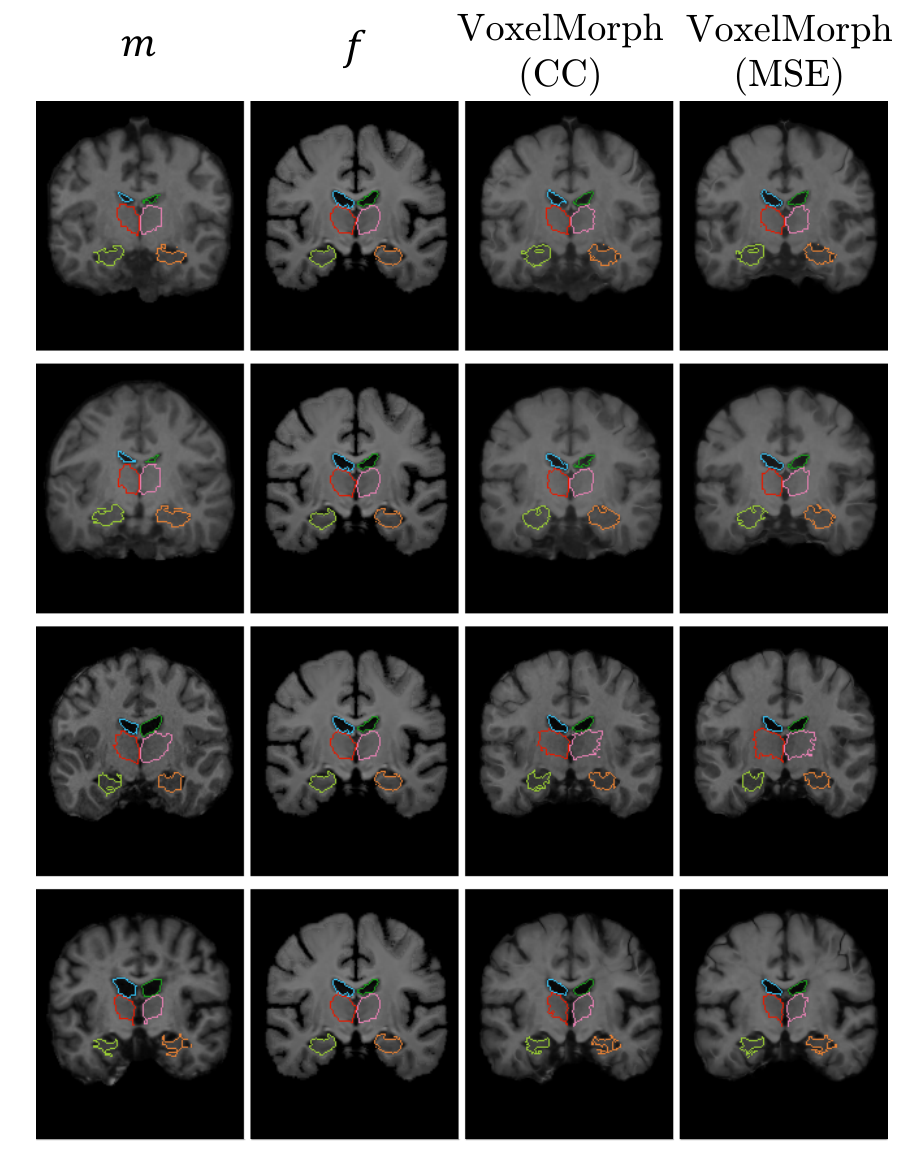}
	\end{center}
	\vspace{-0.3cm}
	\caption{Example MR coronal slices extracted from input pairs (columns 1-2), and resulting $\bmoving \circ \biphi$ for \algname{} {\color{bluedec}using different loss functions}. We overlaid boundaries of a few structures: ventricles (blue/dark green), thalami (red/pink), and hippocampi (light green/orange). A good registration will cause structures in \mbox{$\bmoving \circ \biphi$} to look similar to structures in $\bfixed$. Our models are able to handle various changes in shape of structures, including expansion/shrinkage of the ventricles in rows 2 and 3, and stretching of the hippocampi in row 4. \vspace{-0.4cm}}
	\label{fig:reg_examples}
\end{figure}

\vspace{-0.2cm}
\section{Experiments}
\label{sec:brain-experiments}
{\color{blue}We demonstrate our method on the task of brain MRI registration. We first {\color{bluedec}(Section \ref{sec:main-expt})} present a series of atlas-based registration experiments, in which we compute a registration field between an atlas, or reference volume, and each volume in our dataset. } Atlas-based registration is a common formulation in population analysis, where inter-subject registration is a core problem. The atlas represents a reference, or average volume, and is usually constructed by jointly and repeatedly aligning a dataset of brain MR volumes and averaging them together~\cite{de2004multi}. We use an atlas computed using an external dataset~\cite{fischl2012,sridharan2013}. Each input volume pair consists of the atlas (image $\bfixed$) and a volume from the dataset (image $\bmoving$). Fig.~\ref{fig:reg_examples} shows example image pairs using the same fixed atlas for all examples. {\color{blue} In a second experiment {\color{bluedec}(Section \ref{sec:reganalysis})}, we perform hyper-parameter sensitivity analysis. In a third experiment {\color{bluedec}(Section \ref{sec:amortized})}, we study the effect of training set size on registration, and demonstrate instance-specific optimization. In the fourth experiment {\color{bluedec}(Section \ref{sec:manual})} we present results on a dataset that contains \textit{manual} segmentations. In the next experiment {\color{bluedec}(Section \ref{sec:s2s})}, we train VoxelMorph using random pairs of training subjects as input, and test registration between pairs of unseen test subjects. Finally {\color{bluedec}(Section \ref{sec:auxiliary})}, we present an empirical analysis of registration with auxiliary segmentation data.}  All figures that depict brains in this paper show 2D slices, but all registration is done in 3D.

\subsection{Experimental Setup}

{\color{blue}
	\begin{table*}[tb!]
		\small
		\centering
		\begin{tabular}{>{\color{blue}}c >{\color{blue}}c | >{\color{blue}}c >{\color{blue}}c | >{\color{blue}}c | >{\color{bluedec}}c}
			\textbf{Method}&\textbf{Dice}&\textbf{GPU sec}&\textbf{CPU sec} & $|J_{\bphi}| \le 0$ & $ \% \text{ of }|J_{\bphi}| \le 0$\\
			\hline
			\hline
			\rule{0pt}{1.1em}    
			Affine only&0.584 (0.157)&0&0 & 0 & 0\\
			ANTs SyN (CC) &0.749 (0.136)&-&9059 (2023) & 9662 (6258) & 0.140 (0.091)\\
			NiftyReg (CC) &0.755 (0.143)&-&2347 (202) & 41251 (14336)&0.600 (0.208)\\
			\hline
			\rule{0pt}{1.1em}    
			\algname{} (CC)&0.753 (0.145)&0.45 (0.01)&57 (1) & 19077 (5928) &0.366 (0.114) \\
			\algname{} (MSE)&0.752 (0.140)&0.45 (0.01)&57 (1) & 9606 (4516) &0.184 (0.087)\\
			\hline
		\end{tabular}
		\caption{{\color{blue}Average Dice scores and runtime results for affine alignment, ANTs, {\color{blue}NiftyReg} and \algname{} for the first experiment}.  Standard deviations across structures and subjects are in parentheses. The average Dice score is computed over all structures and subjects. Timing is computed after preprocessing. Our networks yield comparable results to ANTs {\color{blue}and NiftyReg} in Dice score, while operating orders of magnitude faster during testing. We also show the number and percentage of voxels with a non-positive Jacobian determinant for each method, {{\color{bluedec}for our volumes with 5.2 million voxels within the brain}. All methods exhibit less than 1 percent such voxels.} \vspace{-0.25cm}}
		\label{tbl:results}
	\end{table*}   
	\normalsize
}

\begin{figure*}[tb!]
	\begin{center}
		\includegraphics[width=\textwidth]{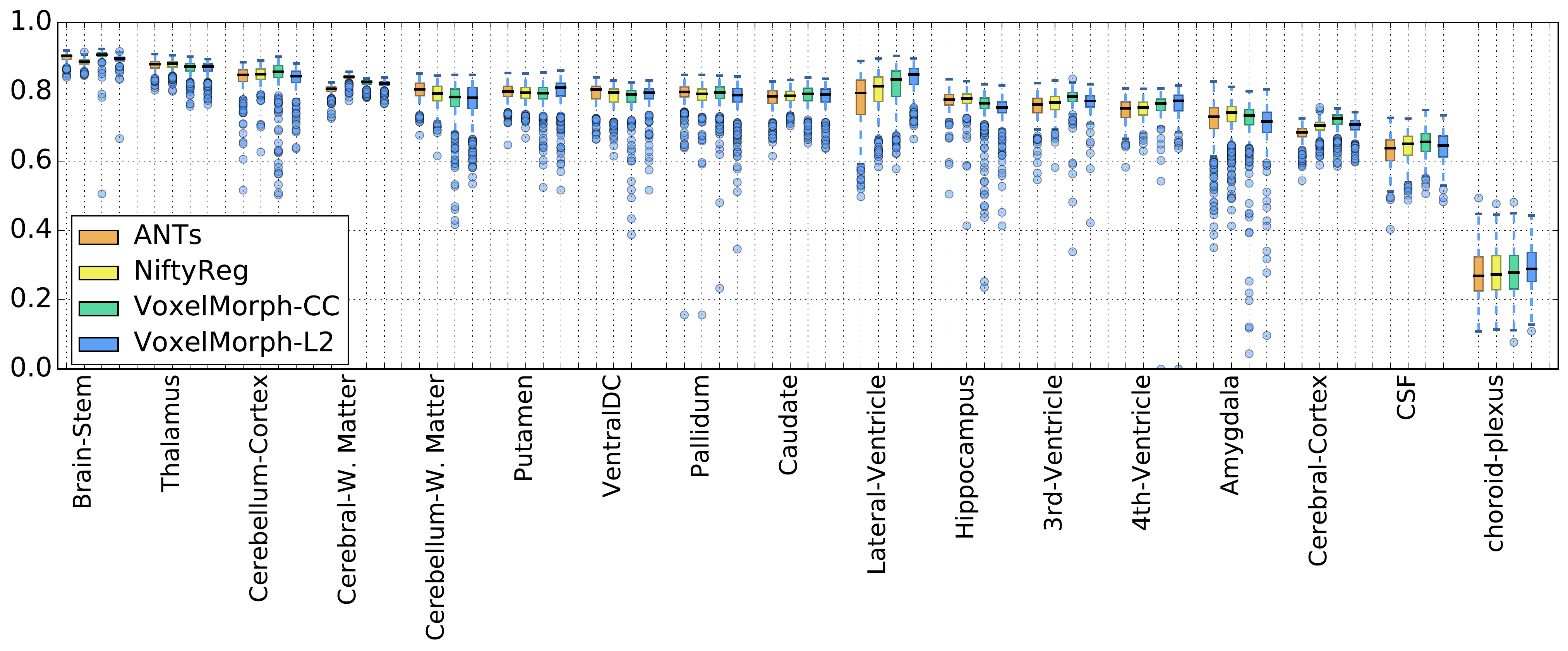}
	\end{center}
	\vspace{-0.25cm}
	\caption{{\color{blue}Boxplots of Dice scores for various anatomical structures for ANTs, {\color{blue}NiftyReg}, and VoxelMorph results for the first {\color{bluedec}(unsupervised)} experiment}. We average Dice scores of the left and right brain hemispheres into one score for this visualization. Structures are ordered by average ANTs Dice score. \vspace{-0.2cm}}
	\label{fig:boxplot}
\end{figure*}

\subsubsection{Dataset}
We use a large-scale, multi-site, multi-study dataset of 3731 T1--weighted brain MRI scans from eight publicly available datasets: %ADNI~\cite{mueller2005ways},
OASIS~\cite{marcus2007open}, ABIDE~\cite{di2014autism}, ADHD200~\cite{milham2012adhd}, MCIC~\cite{gollub2013mcic}, PPMI~\cite{marek2011parkinson}, HABS~\cite{dagley2015harvard}, Harvard GSP~\cite{holmes2015brain}, {\color{blue}and the FreeSurfer Buckner40~\cite{fischl2012}}. Acquisition details, subject age ranges and health conditions are different for each dataset. All scans were resampled to a $256\times 256 \times 256$ grid with $1$mm isotropic voxels. {\color{blue}We carry out standard pre‐processing steps, including affine spatial normalization and brain extraction for each scan using FreeSurfer~\cite{fischl2012}, and crop the resulting images to $160\times 192 \times 224$.} All MRIs were anatomically segmented with FreeSurfer, and we applied quality control using visual inspection to catch gross errors in segmentation results {\color{blue}and affine alignment}. We include all anatomical structures that are at least~$100$ voxels in volume for all test subjects, resulting in 30 structures. We use the resulting segmentation maps in evaluating our registration as described below. We split our dataset into 3231, 250, and 250 volumes for train, validation, and test sets respectively, although we highlight that we do not use any supervised information at any stage. {\color{bluedec} In addition, the Buckner40 dataset is only used for testing, using manual segmentations.}

\subsubsection{Evaluation Metrics}
Obtaining dense \emph{ground truth} registration for these data is not well-defined since many registration fields can yield similar looking warped images. We {\color{bluedec}first} evaluate our method using volume overlap of anatomical segmentations. If a registration field $\bphi$ represents accurate correspondences, the regions in $\bfixed$ and $\bmoving \circ \biphi$ corresponding to the same anatomical structure should overlap well (see Fig.~\ref{fig:reg_examples} for examples). We quantify the volume overlap between structures using the Dice score~\eqref{eqn:dice}.  {\color{blue} We also evaluate the regularity of the deformation fields. Specifically, the Jacobian matrix \mbox{$J_{\phi}(\bp) = \nabla \bphi (\bp) \in \mathcal{R}^{3\times 3}$} captures the local properties of $\bphi$ around voxel $\bp$. {\color{bluedec}We count all non-background voxels for which~$|J_{\phi}(\bp)| \le 0$, where the deformation is not diffeomorphic~\cite{ashburner2007}}}.

\subsubsection{Baseline Methods}
We use Symmetric Normalization (SyN)~\cite{avants2008}, the top-performing registration algorithm in a comparative study~\cite{klein2009} as a first baseline. We use the SyN implementation in the publicly available {\color{blue}Advanced Normalization Tools (ANTs)} software package~\cite{avants2011}, with a cross-correlation similarity measure. Throughout our work with medical images, we found the default ANTs smoothness parameters to be sub-optimal for applying ANTs to our data. {\color{blue} We obtained improved parameters using a wide parameter sweep across multiple datasets, and use those in these experiments. Specifically, we use SyN step size of 0.25, Gaussian parameters (9, 0.2), at three scales with at most 201 iterations each.} {\color{blue}We also use the NiftyReg package, as a second baseline. Unfortunately, a GPU implementation is not currently available, and instead we build a multi-threaded CPU version\footnote{\color{blue}We used the latest source code, updated March, 2018 (tree [4e4525]).}. We searched through various parameter settings to obtain improved parameters, and use the CC cost function, grid spacing of 5, and 500 iterations.}

\subsubsection{VoxelMorph Implementation}
We implemented our method using Keras~\cite{chollet2015} with a Tensorflow backend~\cite{abadi2016}. We extended the 2D linear interpolation spatial transformer layer to {\color{bluedec} $n$-D, and here use $n=3$}. We use the ADAM optimizer~\cite{kingma2014} with a learning rate of $10^{-4}$. {\color{blue}  While our implementation allows for mini-batch stochastic gradient descent, in our experiments each training batch consists of one pair of volumes.}  {\color{blue} Our implementation includes a default of 150,000 iterations. %, however we find that for many instances in our experiments, the network converges by 100,000 or fewer, and we can employ early stopping.
}
 Our code and model parameters are available online at \url{http://voxelmorph.csail.mit.edu}.

\hangindent=3em
\begin{figure}[tb!]
	\begin{center}
		\includegraphics[width=1\linewidth]{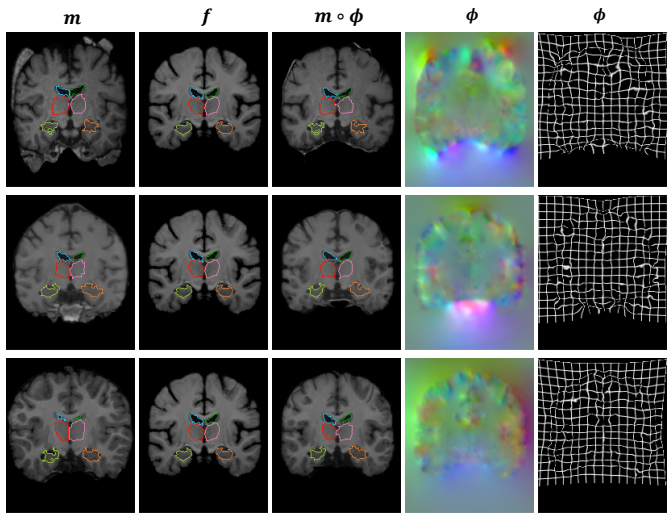}
	\end{center}
	\caption{\color{blue}Example {\color{bluedec}deformation} fields $\bphi$ (columns 4-5) extracted by registering the moving image (column 1) to the fixed image (column 2) {\color{bluedec} in the unsupervised experiment (Section \ref{sec:main-expt}) }. The warped volume $\bmoving \circ \biphi$ is shown in column 3. {\color{bluedec}Displacement} in each spatial dimension is mapped to each of the RGB color channels in column 4. The {\color{bluedec}deformation} fields produced by VoxelMorph (MSE) are smooth within the brain, even when registering moving images that are significantly different from the fixed image.
	}
	\label{fig:brains_wflows}
\end{figure}

\begin{figure}[b!]
	\vspace{-0.25cm}
	\begin{center}
		\includegraphics[width=\linewidth]{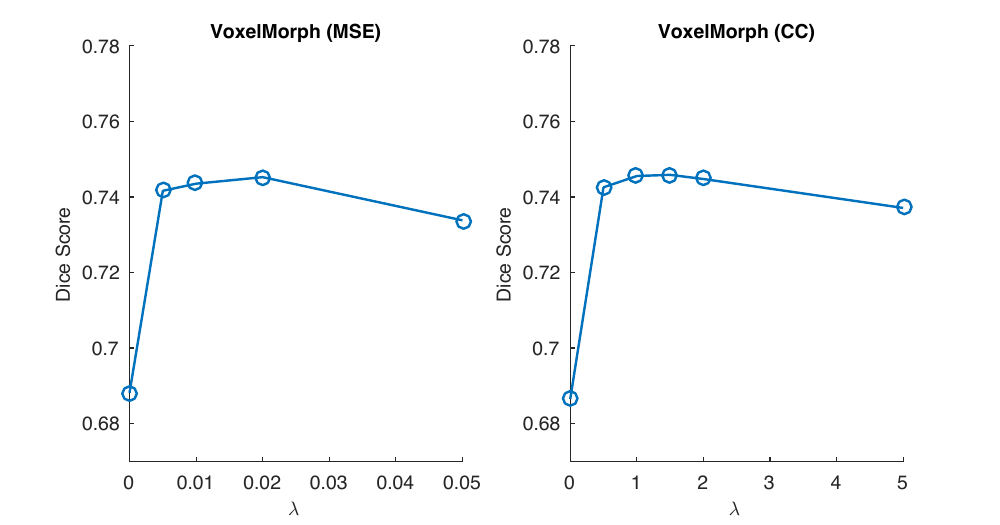}
	\end{center}
	\vspace{-0.3cm}
	\caption{Dice score of validation data for VoxelMorph  with varied regularization parameter $\lambda$. 
		%Shown for comparison in a dashed line is the Dice score for results from ANTs with optimal parameters.
	}
	\label{fig:auc}
\end{figure}

\subsection{Atlas-based Registration}
\label{sec:main-expt}

{\color{blue} In this experiment, we train \algname{} for atlas-based registration. } {\color{bluedec}We train separate \algname{} networks with different~$\lambda$ regularization parameters. We then select the network that optimizes Dice score on our validation set, and report results on our test set.}

Table~\ref{tbl:results} presents average Dice scores computed for all subjects and structures for baselines of only global affine alignment, ANTs, {\color{blue}and NiftyReg, as well as }\algname{} with different losses. \algname{} variants perform comparably to ANTs {\color{blue}and NiftyReg} in terms of Dice\footnote{\color{blue}Both VoxelMorph variants are different from ANTs with paired t-test p-values of $0.003$ and~$0.008$ and with slightly higher Dice values. There is no difference between VoxelMorph (CC) and NiftyReg (p-value of~$0.21$), and no significant difference between VoxelMorph (CC) and VoxelMorph (MSE) (p-value of~$0.09$)}, and are significantly better than affine alignment. Example visual results of the warped images from our algorithms are shown in Figs.~\ref{fig:reg_examples} and~\ref{fig:brains_wflows}. \algname{} is able to handle significant shape changes for various structures.

\begin{figure}[b!]
	\vspace{-0.25cm}
	\begin{center}
		\includegraphics[width=\linewidth]{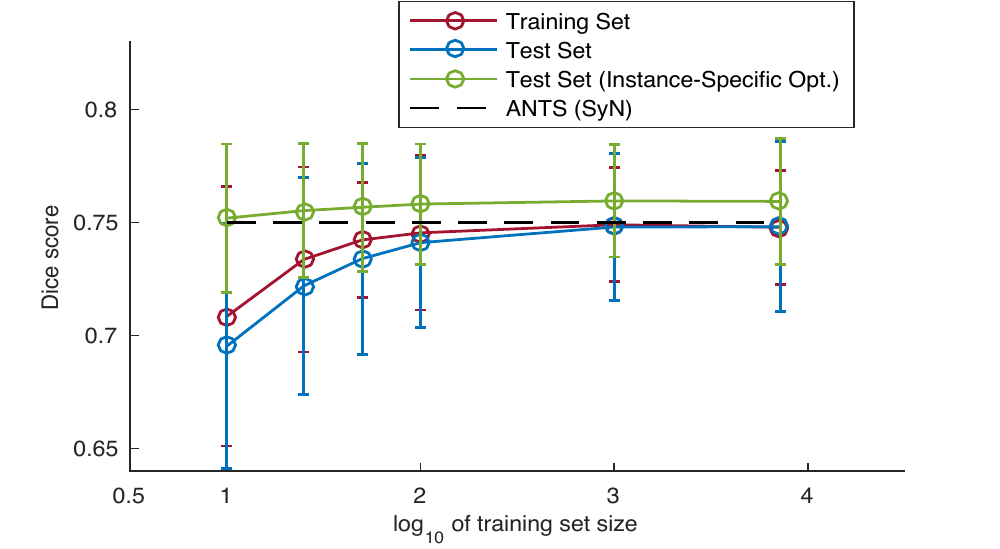}
	\end{center}
	\vspace{-0.3cm}
	\caption{\color{blue} Effect of training set size on accuracy. Also shown are results of instance-specific optimization of deformations, after these are initialized with VoxelMorph outputs using the optimal global parameters resulting from the training phase.}
	\label{fig:amortized}
\end{figure}

Fig.~\ref{fig:boxplot} presents the Dice scores for each structure as a boxplot. For ease of visualization, we average Dice scores of the same structures from the two hemispheres into one score, e.g., the left and right hippocampi scores are averaged. The \algname{} models achieve comparable Dice measures to ANTs {\color{blue}and NiftyReg} for all structures, performing slightly better on some structures such as {\color{blue}the lateral ventricles}, and worse on others such as the hippocampi.

{\color{blue}Table~\ref{tbl:results} includes a count of voxels for which the Jacobian determinant is non-positive. We find that all methods result in deformations with small islands of such voxels, but are diffeomorphic at the vast majority  of voxels (99.4\% - 99.9\%). Figs.~\ref{fig:brains_wflows} and Fig.~\ref{fig:sup-brains} in the supplemental material illustrate several example VoxelMorph deformation fields. VoxelMorph has no explicit constraint for diffeomorphic deformations, {\color{bluedec}but in this setting} the smoothness loss leads to generally smooth and well-behaved results.  ANTs and NiftyReg include implementations that can enforce or strongly encourage diffeomorphic deformations, but during our parameter search these negatively affected runtime or results. {\color{bluedec}In this work, we ran the baseline implementations with configurations that yielded the best Dice scores, which also turned out to produce good deformation regularity.}
%For the purposes of this manuscript, we focused on optimizing these methods for Dice score rather than deformation regularity. 
}

\subsubsection{Runtime}
Table~\ref{tbl:results} presents runtime results using an Intel Xeon (E5-2680) CPU, and a NVIDIA TitanX GPU. We report the elapsed time for computations following the affine alignment preprocessing step, which all of the presented methods share, and requires just a few minutes even on a CPU. ANTs requires two or more hours on the CPU, {\color{blue}while NiftyReg requires {\color{bluedec}roughly} 39 minutes {\color{bluedec} for the given setting}}. ANTs runtimes vary widely, as its convergence depends on the difficulty of the alignment task. Registering two images with \algname{} is, on average, $150$ times faster on the CPU compared to ANTs, and 40 times faster than NiftyReg. When using the GPU, \algname{} computes a registration in under a second. To our knowledge, there is no publicly available ANTs implementation for GPUs. It is likely that the SyN algorithm would benefit from a GPU implementation, but the main advantage of \algname{} comes from not requiring an optimization on each test pair, as can be seen in the CPU comparison. {\color{blue}Unfortunately, the NiftyReg GPU version is unavailable in the current source code on all available repository history.}

\subsection{Regularization Analysis}
\label{sec:reganalysis}
Fig.~\ref{fig:auc} shows average Dice scores for the validation set for different values of the {\color{bluedec}smoothness} regularization parameter $\lambda$. 
%Shown for comparison in a dashed line is the Dice score for alignment with ANTs with optimal parameter values. 
The results vary smoothly over a large range of $\lambda$ values, illustrating that our model is robust to choice of $\lambda$. Interestingly, even setting $\lambda = 0$, which enforces no {\color{bluedec}explicit} regularization on registration, results in a significant improvement over affine registration. This is likely because the optimal network parameters $\theta$ need to register all pairs in the training set well, yielding an implicit dataset regularization for the function $g_{\theta}(\cdot, \cdot)$.

\subsection{Training Set Size and Instance-Specific Optimization}
\label{sec:amortized}
We evaluate the effect of training set size on accuracy, and the relationship between amortized and instance-specific optimization. {\color{blue}Because MSE and CC performed similarly for atlas-based registration, in this section we use MSE. } We train VoxelMorph on subsets of different sizes from our training dataset, and report Dice scores on: (1) the training subset, (2) the {\color{bluedec}held out} test set, and (3) the test set when each deformation is further individually optimized for each test image pair. {\color{bluedec} We perform (3) by “fine-tuning” the displacements~$\bu$ obtained from \algname{} using gradient descent for 100 iterations on each test pair, which took $23.7\pm0.4$ seconds on the GPU or $628.0\pm4.2$ seconds on a single-threaded CPU.}

\begin{figure*}[t!]
	\begin{center}
		\includegraphics[width=0.9\linewidth]{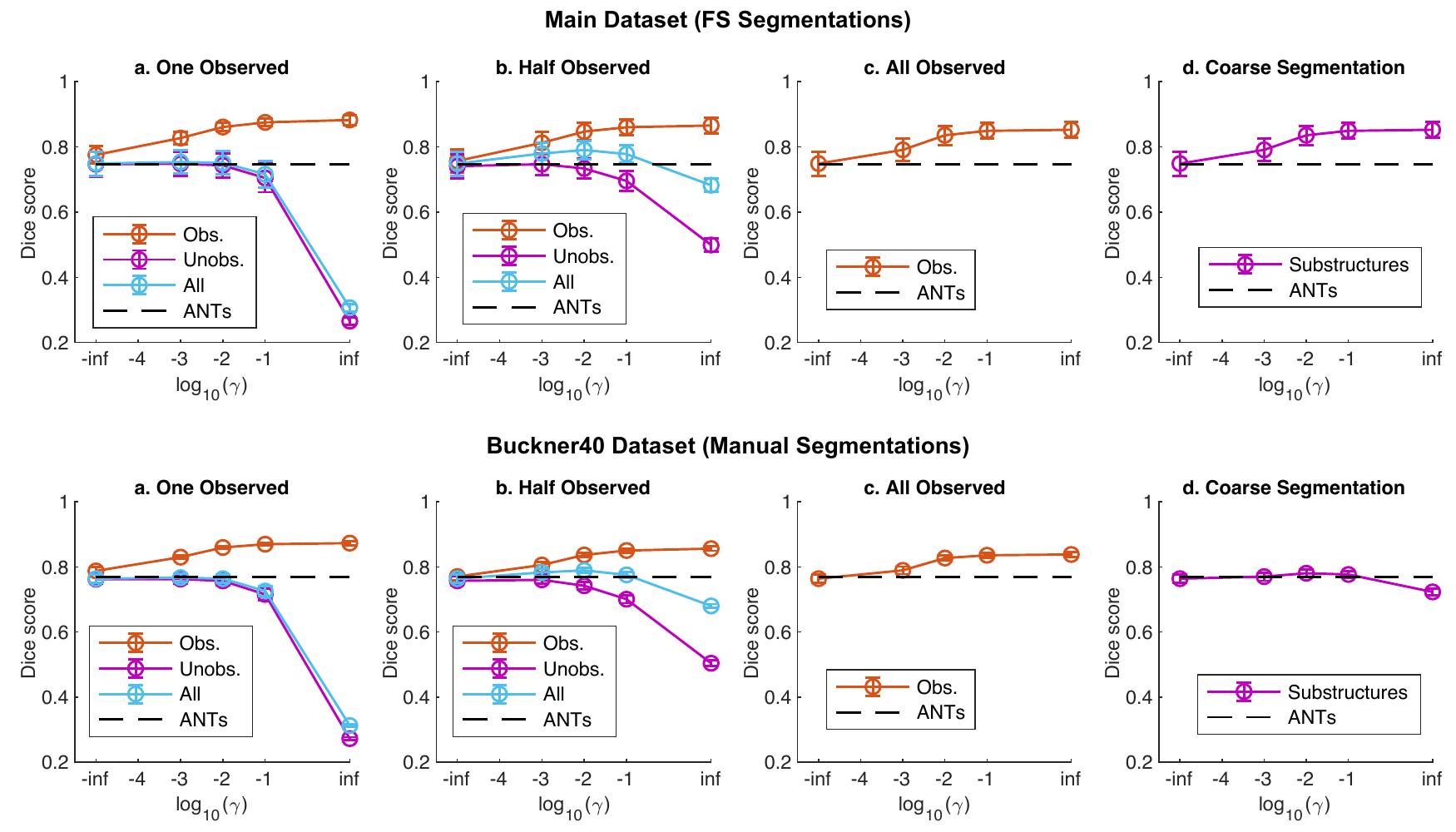}
	\end{center}
	\caption{\color{bluedec}Results on test scans when using auxiliary data during training. Top: testing on the FreeSurfer segmentation of the general test set. Bottom: testing the same models on the manual segmentation of the Buckner40 test set. We test having varying number of observed labels (a-c),
		and having coarser segmentation maps (d). Error bars indicate standard deviations across subjects. The leftmost datapoint in
		each graph for all labels, corresponding to~$\gamma = 0$, indicates results of VoxelMorph without using auxiliary data (unsupervised). $\gamma = \infty$ is achieved by setting the image and smoothness terms to 0.
		We show Dice scores for results from ANTs with optimal parameters, which does not use segmentation maps, for comparison.	\vspace{-0.5cm}}
	\label{fig:supervised}
\end{figure*}

Fig.~\ref{fig:amortized} presents our results. A small training set size of 10 scans results in slightly lower train and test Dice scores compared to larger training set sizes. However, there is no significant difference in Dice scores when training with 100 scans or the full dataset. %These results indicate that training with just~$10$ subjects already yields near-optimal test performance, and training with~$100$ subjects results in state-of-the -art results. 
Further optimizing the \algname{} parameters on each test image pair results in better test Dice scores regardless of training set size, {\color{bluedec}comparable to the state-of-the-art}.

\begin{table}[tb!]
	\small
	\centering
	\begin{tabular}{>{\color{blue}}c >{\color{blue}}c}
		\textbf{Method}&\textbf{Dice}\\
		\hline
		\hline
		\rule{0pt}{1.1em}    
		Affine only&0.608 (0.175)\\
		ANTs SyN (CC) &0.776 (0.130)\\
		NiftyReg (CC) &0.776 (0.132)\\
		\hline
		\rule{0pt}{1.1em}    
		\algname{} (MSE)&0.766 (0.133)\\
		\algname{} (MSE) inst. & 0.776 (0.132)\\
		\algname{} (CC)&0.774 (0.133)\\
		\algname{} (CC) inst. & 0.786 (0.132)\\
		\arrayrulecolor{black}
		\hline
	\end{tabular}
	\caption{\color{blue}Results for manual annotation experiment. We show affine, ANTs, NiftyReg, and \algname{}, where ``inst." indicates additional instance-specific optimization, as described in \textit{Section~\ref{sec:amortized}}.  The average Dice score is computed over all structures and subjects, with standard deviations across structures and subjects in parentheses. }
	\label{tbl:results-buckner40}
\end{table}

\subsection{Manual Anatomical Delineations}
\label{sec:manual}

{\color{blue} Since manual segmentations are not available for most datasets, the availability of FreeSurfer segmentations enabled the broad range of experiments above. In this experiment, we use VoxelMorph models already trained in \textit{Section~\ref{sec:main-expt}} to test registration on the (unseen) Buckner40 dataset containing~$39$ scans. This dataset contains expert manual delineations of the same anatomical structures used in previous experiments, which we use here for evaluation. We also compute VoxelMorph with instance-specific optimization, as described in \textit{Section~\ref{sec:amortized}}. The Dice score results, shown in Table~\ref{tbl:results-buckner40}, show that VoxelMorph using cross-correlation loss behaves comparably to ANTs and NiftyReg using the same cost function, consistent with the first experiment where we evaluated on FreeSurfer segmentations.  VoxelMorph with instance-specific optimization further improves the results, similar to the previous experiment. On this dataset, results using VoxelMorph with MSE loss obtain slightly lower scores, but are improved by the instance-specific optimization procedure to be comparable to ANTs and NiftyReg. }

%\vspace{0.5cm}  % otherwise title formatted funnily in current version
\subsection{\color{blue}Subject-to-Subject Registration}
\label{sec:s2s}

{\color{blue} In this experiment, we train \algname{} for subject-to-subject registration. Since there is more variability in each registration, we double the number of features for each network layer. We also compute VoxelMorph with instance-specific optimization, as described in \textit{Section~\ref{sec:amortized}}. Table~\ref{tbl:results-s2s} presents average test Dice scores {\color{bluedec}on 250 randomly selected test pairs for registration}. Consistent with literature, we find that the normalized cross correlation loss leads to more robust results compared to using the MSE loss. VoxelMorph {\color{bluedec}(with doubled feature counts)} Dice scores are comparable with ANTs and slightly below NiftyReg, while results from VoxelMorph with instance-specific optimization are comparable to both baselines.}

\begin{table}[t!]
	\small
	\centering
	\begin{tabular}{>{\color{bluedec}}c >{\color{bluedec}}c}
		\textbf{Method}&\textbf{Dice}\\
		\hline
		\hline
		\rule{0pt}{1.1em}    
		Affine only&0.579 (0.173)\\
		ANTs SyN (CC) &0.761 (0.117)\\
		NiftyReg (CC) &0.772 (0.117)\\
		\hline
		\rule{0pt}{1.1em}   
		\algname{} (MSE)&0.727 (0.146)\\
		\algname{} x2 (MSE)&0.750 (0.058)\\ \vspace{0.1cm}
		\algname{} x2 (MSE) inst. & 0.764 (0.048)\\
		\algname{} (CC)&0.737 (0.139)\\ 
		\algname{} x2 (CC) &0.763 (0.049)\\	 
		\algname{} x2 (CC) inst. & 0.772 (0.119) \\  
		\hline
	\end{tabular}
	\caption{\color{bluedec} Results for subject-to-subject alignment using affine, ANTs, and \algname{} variants, where ``x$2$'' refers to a model where we doubled the number of features to account for the increased inherent variability of the task, and ``inst." indicates additional instance-specific optimization. }
	\label{tbl:results-s2s}
\end{table}

\subsection{Registration with Auxiliary Data}
\label{sec:auxiliary}
In this section, we evaluate \algname{} {\color{bluedec} when using  segmentation maps during training with loss function~\eqref{eqn:semi}}. {\color{blue}Because MSE and CC performed similarly for atlas-based registration, in this section we use {\color{bluedec}MSE with~$\lambda=0.02$}. } We present an evaluation of our model in two practical scenarios: (1) when subsets of anatomical structure labels are available during training, and (2) when coarse segmentations labels are available during training. We use the same train/validation/test split as the previous experiments.

\begin{figure}[tb!]
	\begin{center}
		\includegraphics[width=1\linewidth]{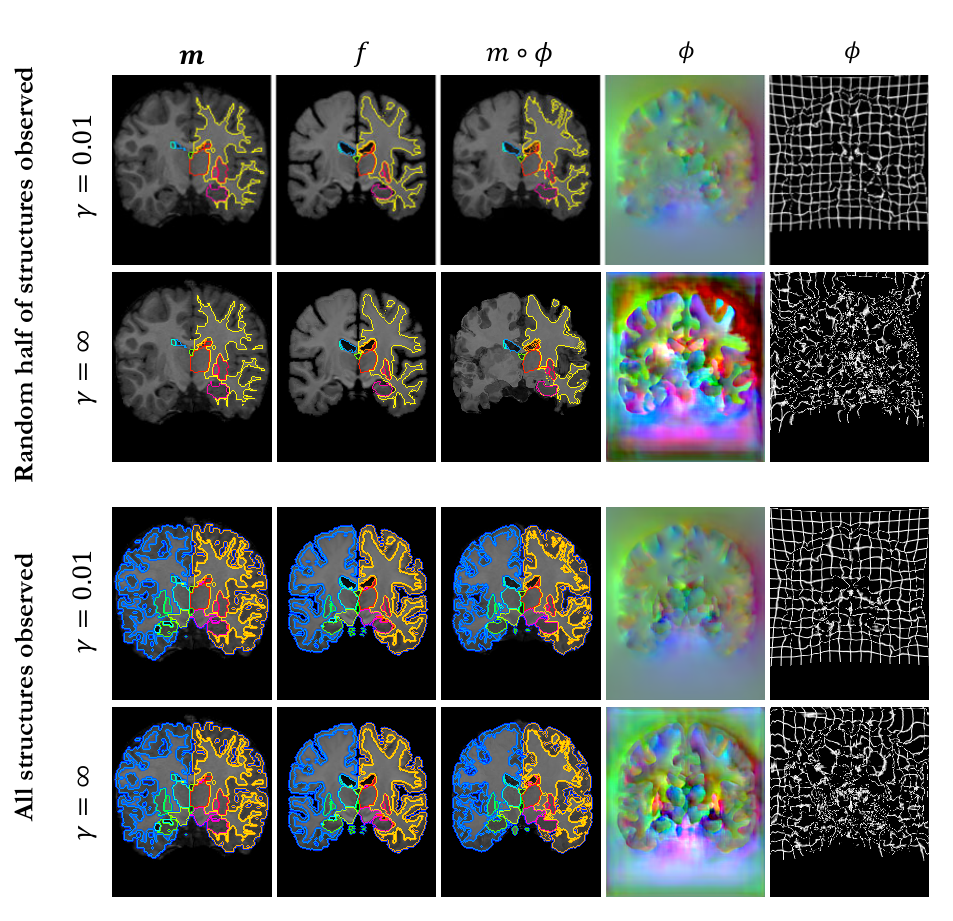}
	\end{center}
	\vspace{-0.25cm}
	\caption{{\color{blue}Effect of $\gamma$ on warped images and deformation fields. We show the moving image, fixed image, and warped image (columns 1-3) with the structures that were observed at train time overlaid. The resulting deformation field is visualized in columns 4 and 5. While providing better Dice scores for observed structures, the deformation fields resulting from training with~$\gamma=\infty$ are far more irregular than those using~$\gamma=0.01$. Similarly, the warped image are visually less coherent for~$\gamma=\infty$. \vspace{-0.2cm}}}
	\label{fig:gamma-examples}
\end{figure}

\subsubsection{Training with a subset of anatomical labels}
In many practical settings, it may be infeasible to obtain training segmentations for all structures. We therefore first consider the case where segmentations are available for only a subset of the 30 structures. We refer to structures present in segmentations as \emph{observed}, and the rest as \emph{unobserved}. {\color{blue} We considered three scenarios, when: one, 15 (half), and 30 (all) structure segmentations are observed. The first two experiments essentially simulate different amounts of partially observed segmentations. For each experiment, we train separate models on different subsets of observed structures, as follows. For single structure segmentations, we manually selected four important structures for four folds (one for each fold) of the experiment: hippocampi, cerebral cortex, cerebral white matter, and ventricles. For the second experiment, we randomly selected 15 of the 30 structures, with a different selection for each of five folds. For each fold and each subset of observed labels, we use the segmentation maps at training, and show results on test pairs where segmentation maps are not used. }

Fig.~\ref{fig:supervised}a-c shows Dice scores for both the observed and unobserved labels when sweeping $\gamma$ in~\eqref{eqn:semi}, the auxiliary regularization trade-off parameter. {\color{bluedec} We train our models with FreeSurfer annotations, and show results on both the general test set using FreeSurfer annotations (top) and the Buckner40 test set with manual annotations (bottom). The extreme values $\gamma =0$ (or $\log \gamma = -\infty$) and $\gamma = \infty$ serve as theoretical extremes, with $\gamma =0$ corresponding to unsupervised \algname{}, and $\gamma = \infty$ corresponding to \algname{} trained \emph{only} with auxiliary labels, without the smoothness and image matching objective terms.}

 In general, \algname{} with auxiliary data significantly outperforms (largest p-value \mbox{$<10^{-9}$} among the four settings) unsupervised \algname{} (equivalent to $\gamma = 0$ or $\log \gamma = - \infty$) and ANTs on observed structures {\color{bluedec}in terms of Dice score}. Dice score on observed labels generally increases with an increase in $\gamma$. 

\begin{table*}[tb!]
	\centering
	\begin{tabular}{>{\color{bluedec}}c | >{\color{bluedec}}c >{\color{bluedec}}c >{\color{bluedec}}c >{\color{bluedec}}c >{\color{bluedec}}c}
		\textbf{Setting}&\textbf{$0$}&\textbf{$0.001$}&\textbf{$0.01$}&\textbf{$0.1$}&\textbf{$\infty$}\\
		\hline
		\hline
		\rule{0pt}{1.1em}    
		one (count)&9606 (4471)& 10435 (4543)& 22998 (3171) & 121546 (12203)& 685811 (6878)\\
		one (\%)&0.18 (0.09)&0.20 (0.09)&0.44 (0.06)&2.33 (0.23)&13.14 (0.13)\\
		half (count)&9606 (4471)&9470 (4008)&17886 (4919)&86319 (13851)& 516384 (7210) \\
		half (\%)&0.18 (0.09)&0.18 (0.08)&0.34 (0.09)&1.65 (0.27)&9.90 (0.14)  \\
		all (count)&9606 (4471)&10824 (5029)&19226 (4471)&102295 (14366)&528552 (8720)	 \\
		all (\%)& 0.18 (0.09)&0.21 (0.10)&0.37 (0.09)&1.96 (0.28)&10.13 (0.17)\\
		coarse (count)&9606 (4471)&9343 (4117)&15190 (4416)&76677 (11612)&564493 (7379) \\
		coarse (\%)&0.18 (0.09)&0.18 (0.08)&0.29 (0.08)&1.47 (0.22)&10.82 (0.14)  \\
		\hline
	\end{tabular}
	\caption{\color{bluedec} Regularity of deformation fields when training with auxiliary segmentations obtained using FreeSurfer, MSE loss function and smoothness parameter of 0.02, measured using count and percentage of the number of voxels with non-positive Jacobian determinants.}
	\label{tbl:supervised-examples}
\end{table*}

Interestingly, \algname{} (trained with auxiliary data) yields improved Dice scores for unobserved structures compared to the unsupervised variant for a range of $\gamma$ values (see Fig.~\ref{fig:supervised}a-b), even though these segmentations were not explicitly observed during training. {\color{blue} When \textit{all} structures that we use during evaluation are observed during training, we find {\color{bluedec}good} Dice results {\color{bluedec}at higher~$\gamma$ values} (Fig~\ref{fig:supervised}c.).  {\color{blue}Registration accuracy for unobserved structures starts declining when~$\gamma$ is large, in the range~$\log\gamma \in [-3, -2]$. This can be interpreted as the range where the model starts to over-fit to the observed structures - that is, it continues to improve the Dice score for observed structures while harming the registration accuracy for the other structures (Fig.~\ref{fig:supervised}c)}

\subsubsection{Training with coarse labels}
We consider the scenario where only coarse labels are available, such as when all the white matter is segmented as one structure. This situation enables evaluation of how the auxiliary data affects anatomical registration at finer scales, within the coarsely delineated structures. To achieve this, we merge the 30 structures into four broad groups: white matter, gray matter, cerebral spinal fluid (CSF) and the brain stem, and evaluate the accuracy of the registration on the original structures.

Fig.~\ref{fig:supervised}d {\color{bluedec}(top)} presents mean Dice scores over the original 30 structures with varying $\gamma$. With $\gamma$ of 0.01, we obtain an average Dice score of {\color{bluedec}$0.78 \pm 0.03$ on FreeSurfer segmentations. This is roughly a $3$ Dice point improvement} over \algname{} without auxiliary information (p-value $< 10^{-10}$).

\subsubsection{\color{bluedec}Regularity of Deformations} 
We also evaluate the regularity of the deformation fields both visually and by computing the number of voxels for which the determinant of the Jacobian is non-positive. {\color{bluedec}Table~\ref{tbl:supervised-examples} provides the quantitative regularity measure  for all $\gamma$ values, showing that VoxelMorph deformation regularity degrades slowly as a function of~$\gamma$ (shown on a log scale), with roughly 0.2\% of the voxels exhibiting folding at the lowest parameter value, and \textit{at most} 2.3\% when $\gamma = 0.1$. Deformations from models that don't encourage smoothness, at the extreme value of~$\gamma=\infty$, exhibit $10$--$13\%$ folding voxels. A lower~$\gamma$ value such as~$\gamma=0.01$ therefore provides a good compromise of high Dice scores for all structures while avoiding highly irregular deformation fields, and avoiding overfitting as described above.} Fig~\ref{fig:gamma-examples} shows examples of deformation fields {\color{bluedec}for ~$\gamma=0.01$ and~$\gamma=\infty$, and we include more figures in the supplemental material for each experimental setting. }

\subsubsection{\color{bluedec}Testing on Manual Segmentation Maps}
{\color{bluedec}
We also test these models on the manual segmentations in the Buckner40 dataset used above, resulting in Fig. ~\ref{fig:supervised} (bottom). We observe a behavior consistent with the conclusions above, with smaller Dice score improvements, possibly due to the higher baseline Dice scores achieved on the Buckner40 data.}

\vspace{-0.2cm}
\section{Discussion and Conclusion}
\algname{} with unsupervised  loss performs comparably to the state-of-the-art ANTs {\color{blue}and NiftyReg} software in terms of Dice score, while reducing the computation time from hours to minutes on a CPU and under a second on a GPU. %{\color{red}Using the MSE loss leads to comparable results in some experiments, or slightly lower Dice scores in other.} 
\algname{} is flexible and handles both partially observed or coarsely delineated auxiliary information during training, {\color{bluedec}which can lead to improvements in Dice score while still preserving the runtime improvement.}

\algname{} performs amortized optimization, learning {\color{bluedec}global} function parameters that are optimal for an entire training dataset. As Fig.~\ref{fig:amortized} shows, the dataset need not be large: with only 100 training images, \algname{} leads to state-of-the-art registration quality scores for both training and test sets. Instance-specific optimization further improves \algname{} performance by one Dice point. This is a small increase, illustrating that amortized optimization {\color{bluedec}can} lead to nearly optimal registration. % Because this increase improves on traditional state-of-the-art algorithm, it is possible that the added regularization of amortized optimization helps \algname{} converge to a better solution than the traditional state-of-the-art algorithms. 

We performed a thorough set of experiments demonstrating that, for a reasonable choice of~$\gamma$, the availability of anatomical segmentations during training significantly improves test registration performance with \algname{} {\color{bluedec}(in terms of Dice score)} {\color{blue}while providing smooth deformations} {\color{bluedec}(e.g. for~$\gamma=0.01$, less than~$0.5\%$ folding voxels)}. The performance gain varies based on the quality and number of anatomical segmentations available. Given a single labeled anatomical structure during training, the accuracy of registration of test subjects \textit{for that label} increases, without negatively impacting other anatomy. If half or all of the labels are observed, or even a coarse segmentation is provided at training, registration accuracy improves for all labels during test.  While we experimented with one type of auxiliary data in this study, \algname{} can leverage other auxiliary data, such as different modalities or anatomical keypoints. {\color{bluedec} Increasing $\gamma$ also increases the number of voxels exhibiting a folding of the registration field. This effect may be alleviated by using a diffeomorphic deformation representation for \algname{}, as introduced in recent work~\cite{dalca2018}}.

\algname{} is a general learning model, and is not limited to a particular image type or anatomy -- it may be useful in other medical image registration applications such as cardiac MR scans or lung CT images. With an appropriate loss function such as mutual information, the model can also perform multimodal registration. \algname{} promises to significantly speed up medical image analysis and processing pipelines, while opening novel directions in learning-based registration.

\ifCLASSOPTIONcaptionsoff
  \newpage
\fi

\IEEEtriggeratref{59}


%\bibliographystyle{IEEEtran}
\bibliographystyle{IEEEtran}
\bibliography{egbib} % IEEEabrv

\appendices


%\vfill


\clearpage
\begin{center}
    {\Large SUPPLEMENTARY MATERIAL}
\end{center}
%\subsection{Network architecture details}
%Here we describe the full network architecture used in our experiments. Let Ck-s denote a 3D convolution layer with $k$ filters and a stride of $s$. All convolutions use a kernel size of $3$, and are followed by a LeakyReLU layer with $\alpha=0.2$. We use the notation $c_i$ to denote the output of each convolution in the encoder. Let U represent a 3D upsampling layer, and A$c_i$ denote concatenation with the volume $c_i$ from the encoder.\\
%
%\noindent Encoder: C16-2 ($c_1$), C32-2 ($c_2$), C32-2 ($c_3$), C32-2 ($c_4$)\\
%Decoder: C32, U, A$c_3$, C32, U, A$c_2$, C32, U, A$c_1$, C32, C32, U, C16, C16
\begin{figure}[b]\setlength{\hfuzz}{1.1\columnwidth}
\hspace{-250pt}
\begin{minipage}{\textwidth}
%\ttfamily 

%\begin{figure*}[h!]
	\begin{center}
		\includegraphics[width=0.7\linewidth]{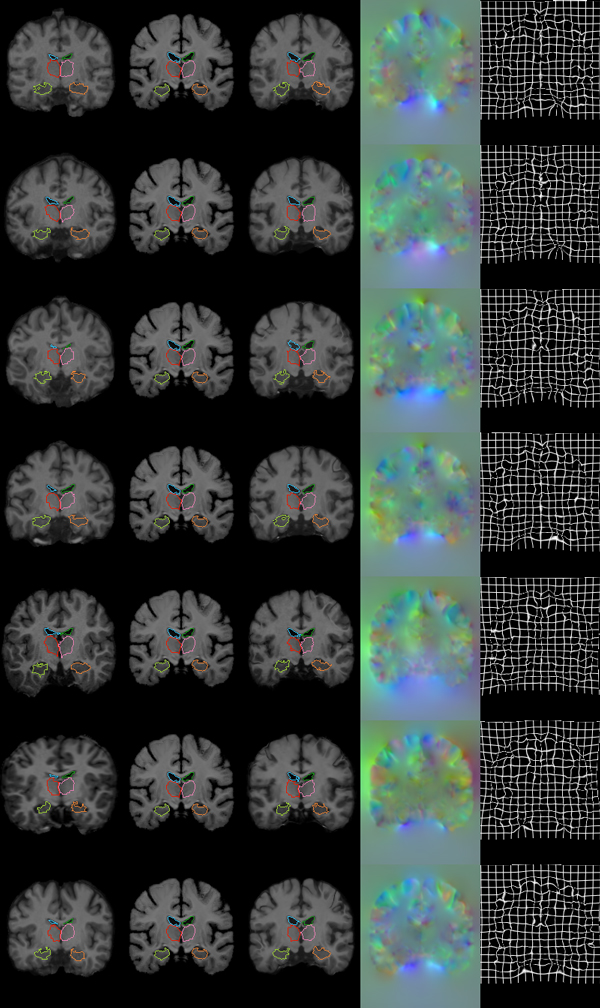}
	\end{center}
	\caption{\color{blue}Example atlas-based VoxelMorph flow fields $\bphi$ (columns 4-5) extracted by registering the moving image (column 1) to the fixed image (column 2). The warped image $\bmoving \circ \biphi$ is shown in column 3. 
	}
	\label{fig:sup-brains}
%\end{figure*}
\end{minipage}
\end{figure}

\begin{figure*}[h!]
	\begin{center}
	\begin{subfigure}[b]{0.33\linewidth}
		\includegraphics[width=\linewidth]{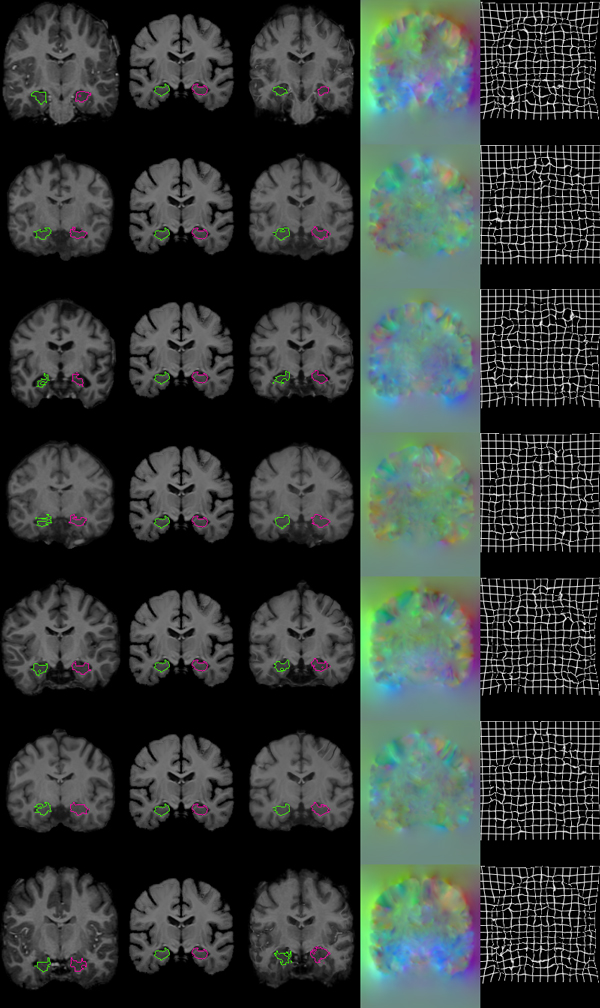}
	\end{subfigure} 
	~
		\begin{subfigure}[b]{0.33\linewidth}
		\includegraphics[width=\linewidth]{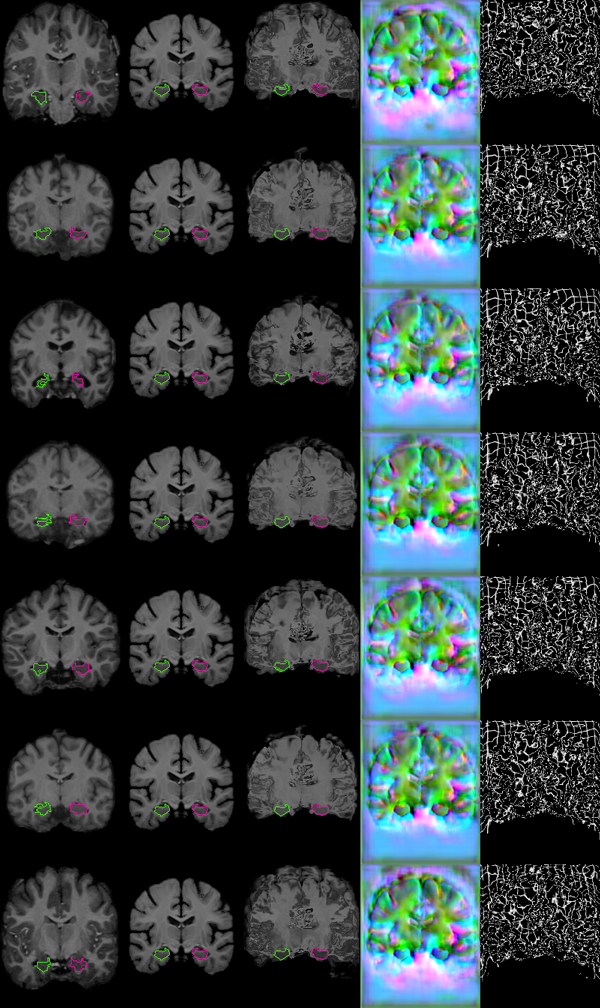}
	\end{subfigure}
		\end{center}
	\caption{\color{blue}Auxiliary data experiment where the left and right hippocampus labels are observed at train time. We show the moving image, fixed image and warped image (columns 1-3) with the observed labels overlaid, and the resulting deformation fields (columns 4-5). We use the optimal $\gamma=0.01$ (left) and the extreme $\gamma=\infty$ (right).  
	}
	\label{fig:supp_labels_hippo}
\end{figure*}

\begin{figure*}[h!]
	\begin{center}
	\begin{subfigure}[b]{0.33\linewidth}
		\includegraphics[width=\linewidth]{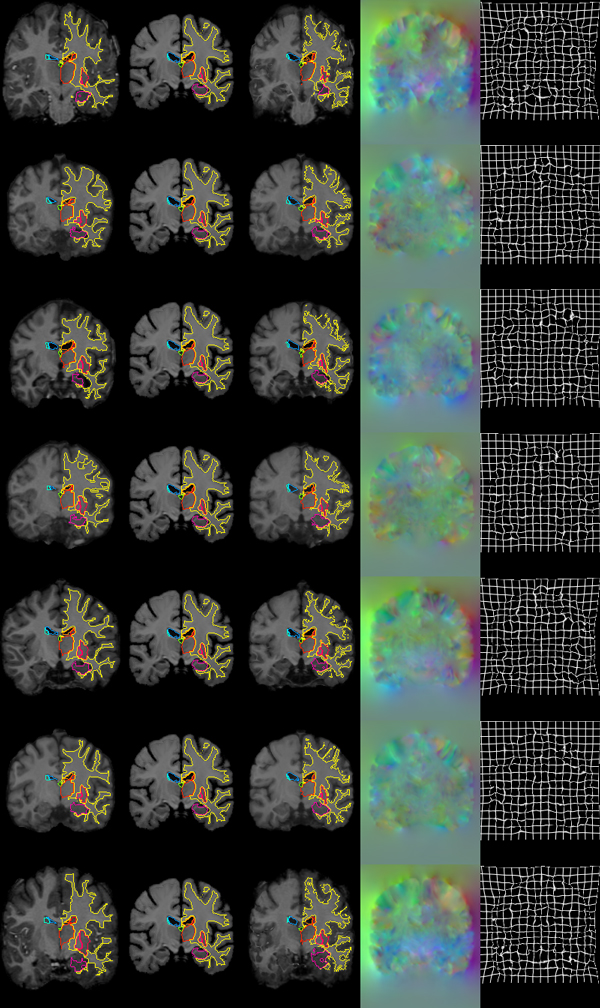}
	\end{subfigure} 
	~
		\begin{subfigure}[b]{0.33\linewidth}
		\includegraphics[width=\linewidth]{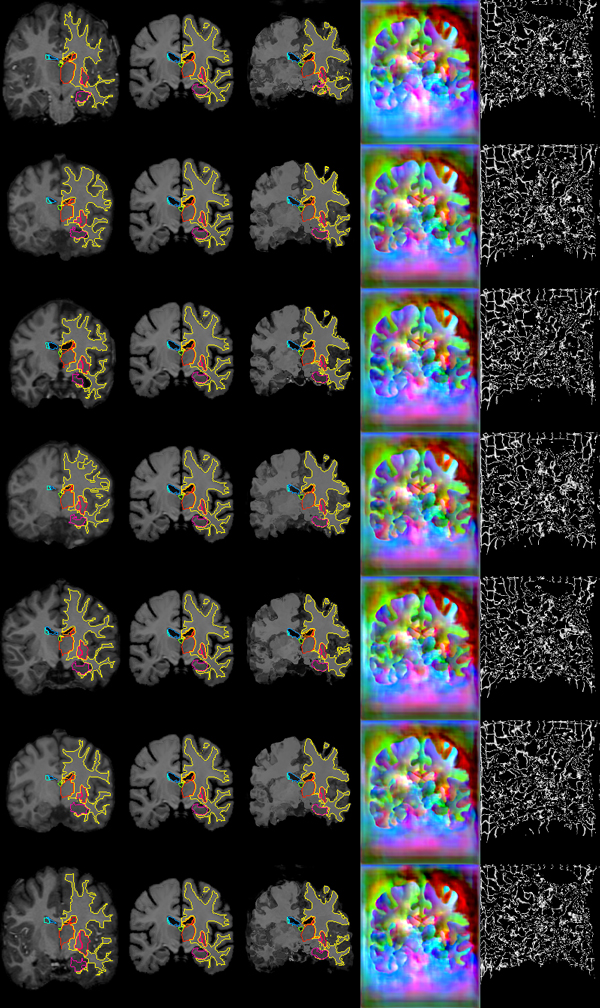}
	\end{subfigure}
		\end{center}
	\caption{\color{blue}Auxiliary data experiment where a random half of the labels are observed at train time. We show the moving image, fixed image and warped image (columns 1-3) with the observed labels overlaid, and the resulting deformation fields (columns 4-5). We use the optimal $\gamma=0.01$ (left) and the extreme $\gamma=\infty$ (right).  
	}
	\label{fig:supp_labels_half}
\end{figure*}

\begin{figure*}[h!]
	\begin{center}
	\begin{subfigure}[b]{0.33\linewidth}
		\includegraphics[width=\linewidth]{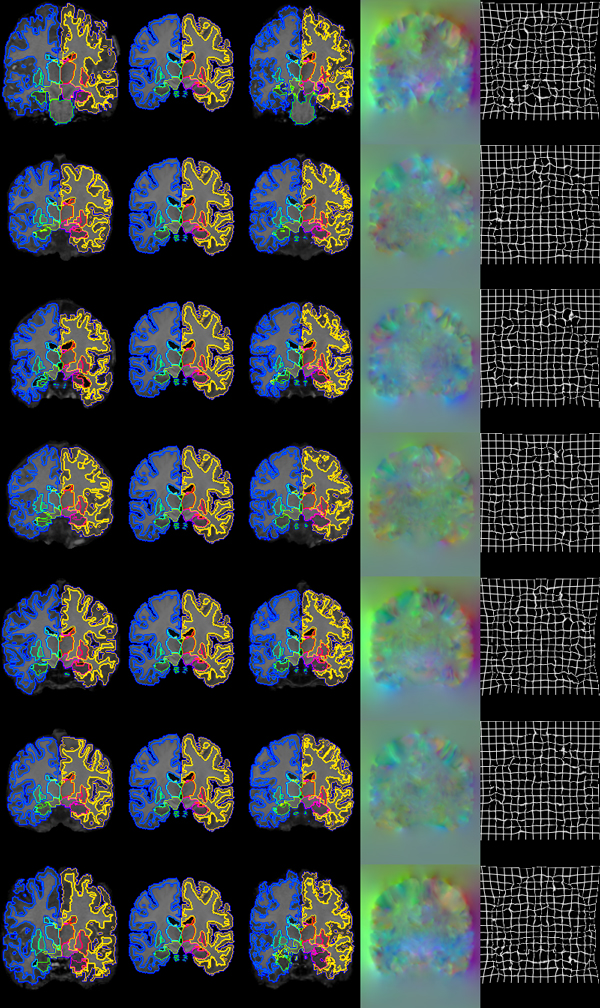}
	\end{subfigure} 
	~
		\begin{subfigure}[b]{0.33\linewidth}
		\includegraphics[width=\linewidth]{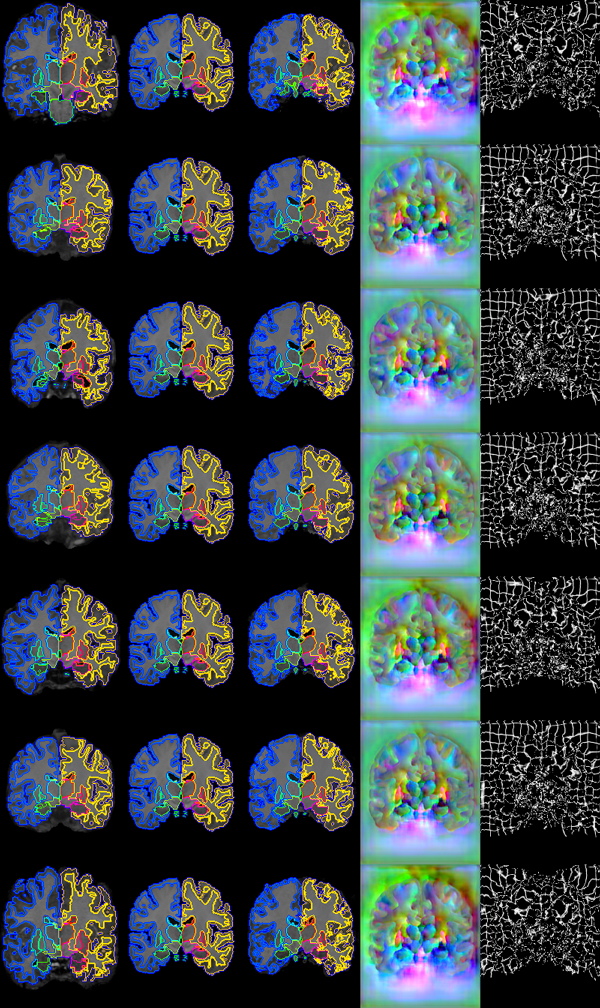}
	\end{subfigure}
		\end{center}
	\caption{\color{blue}Auxiliary data experiment where all labels are observed at train time. We show the moving image, fixed image and warped image (columns 1-3) with the observed labels overlaid, and the resulting deformation fields (columns 4-5). We use the optimal $\gamma=0.01$ (left) and the extreme $\gamma=\infty$ (right).  
	}
	\label{fig:supp_labels_all}
\end{figure*}

\begin{figure*}[h!]
	\begin{center}
	\begin{subfigure}[b]{0.33\linewidth}
		\includegraphics[width=\linewidth]{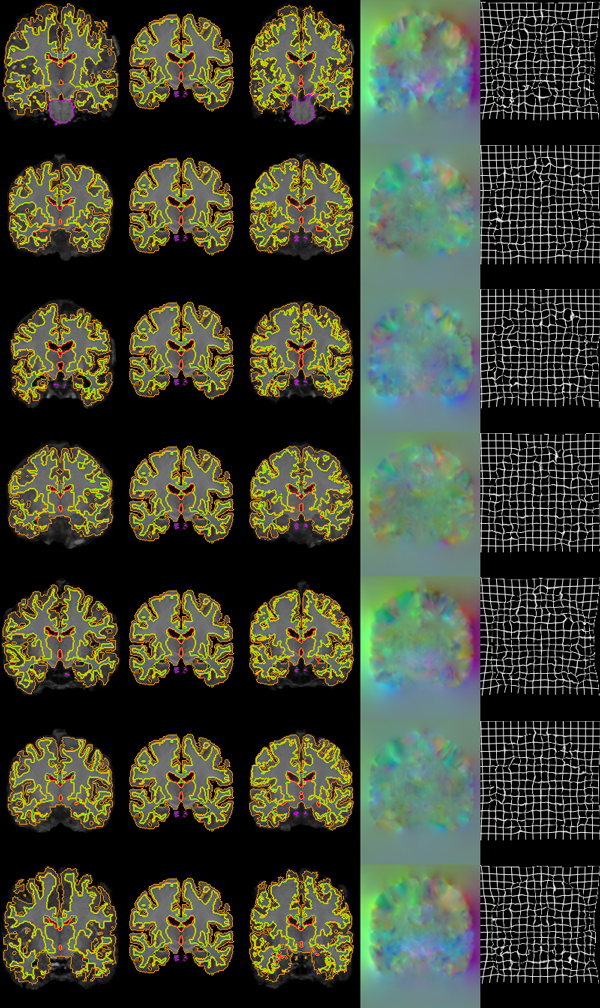}
	\end{subfigure} 
	~
		\begin{subfigure}[b]{0.33\linewidth}
		\includegraphics[width=\linewidth]{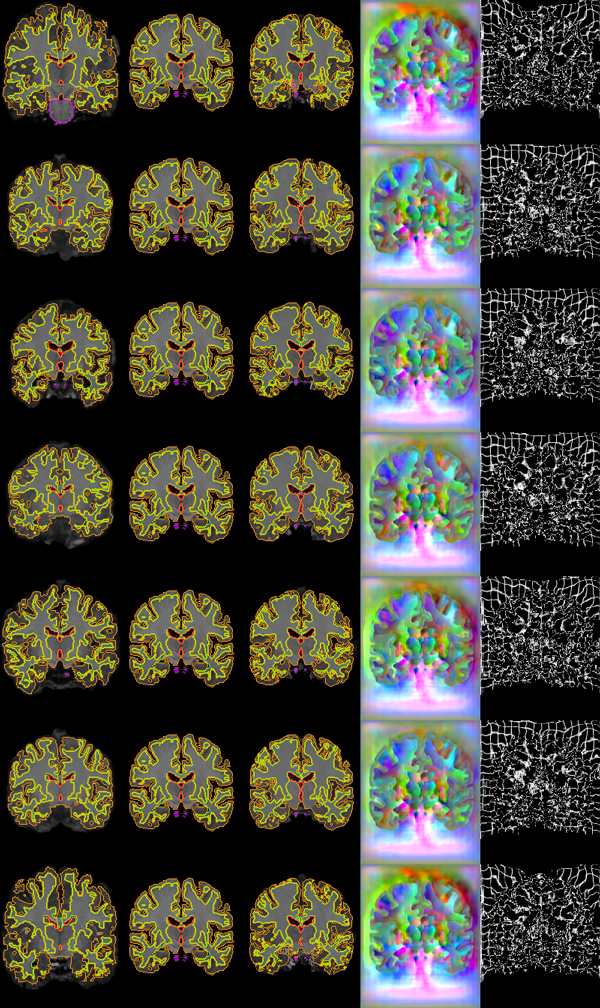}
	\end{subfigure}
		\end{center}
	\caption{\color{blue}Auxiliary data experiment where coarse labels are observed at train time. We show the moving image, fixed image and warped image (columns 1-3) with the observed labels overlaid, and the resulting deformation fields (columns 4-5). We use the optimal $\gamma=0.01$ (left) and the extreme $\gamma=\infty$ (right).  
	}
	\label{fig:supp_labels_coarse}
\end{figure*}

\end{document}